%% file: output.tex
\documentclass[a4paper,12pt,headsepline]{scrartcl}

\usepackage{graphicx}

\usepackage{ngerman}

\usepackage[german, english]{babel} 

\usepackage[right]{eurosym}

\usepackage[utf8]{inputenc}

\usepackage[T1]{fontenc}

\usepackage{lmodern}
\usepackage{fix-cm}

\usepackage{longtable}

\usepackage{colortbl}

\usepackage{array}

\usepackage{geometry}

\usepackage{fancybox}

\usepackage[hyphens,obeyspaces,spaces]{url}

\usepackage{color}

\usepackage{amssymb}

\usepackage[bookmarksnumbered,hyperfootnotes=false]{hyperref}

\usepackage{fancyhdr} 

\usepackage{array}

\usepackage{setspace}

\usepackage{capt-of}

\usepackage{makeidx}

\usepackage{listings}

\usepackage[acronym, toc]{glossaries}

\makeindex

\makeglossaries

\hypersetup{
  pdftitle    = {Automated dataset generation for image recognition using the example of taxonomy},
  pdfauthor   = {Jaro Milan Zink},
  pdfkeywords = {artificial intelligence, automation, image recognition, image classification, big data, taxonomy}
}

\include{definitionen/glossar}

\include{definitionen/variablen}


\begin{document}

\spacing{1.35}

\include{extras/deckblatt}

\include{extras/erklaerung}

\include{extras/abstract}

\newpage
\tableofcontents

\newpage
\fancyhead[L]{List of figures / tables / listings}
\addcontentsline{toc}{section}{List of figures} 
\listoffigures
\addcontentsline{toc}{section}{List of tables}
\listoftables
\addcontentsline{toc}{section}{List of listings}
\lstlistoflistings

\newpage
\fancyhead[L]{Glossary} 
\printglossary

\newpage
\fancyhead[L]{List of abbreviations}
\printglossary[type=\acronymtype]

\newpage
\fancyhead[L]{\nouppercase{\leftmark}} 
\include{kapitel/1_introduction}
\include{kapitel/2_related_work}
\include{kapitel/3_basics}
\include{kapitel/4_requirements_analysis}
\include{kapitel/5_conception}
\include{kapitel/6_implementation}
\include{kapitel/7_evaluation}
\include{kapitel/8_conclusion}

\newpage
\addcontentsline{toc}{section}{References}
\bibliography{definitionen/literatur}

\include{extras/anhang}

\end{document}

%% file: definitionen/glossar.tex
\newglossaryentry{ai_gls}
{
    name=artificial intelligence,
    description={The approach of \acrshort{ai} (often) uses an implementation of the mapping of the structure inside a human brain (a so called "neural network") to enable computers to "think" like a human being}
}
\newglossaryentry{ui_gls}
{
    name=user interface,
    description={An interface for the user to interact with a device. Usually provided in graphic form, like a website or mobile app, e.g.}
}
\newglossaryentry{framework}
{
    name=framework,
    description={A kind of skeleton/library used in programming. Frameworks provide expandable functionality which is used by software developers to produce more straightforward code}
}
\newglossaryentry{ml_gls}
{
    name=machine learning,
    description={Describes the process of creating an \acrshort{ai}. This usually includes the development of the software, a training- and a test-phase}
}
\newglossaryentry{biodiv}
{
    name=biodiversity,
    description={The variety of living organisms in the world-wide biological ecosystem(s)}
}

\newacronym{ai}{AI}{artificial intelligence}
\newacronym{nn}{NN}{neural network}
\newacronym{ml}{ML}{machine learning}
\newacronym{ui}{UI}{user interface}
\newacronym{sql}{SQL}{Structured Query Language}
\newacronym{ilsvrc}{ILSVRC}{ImageNet Large Scale Visual Recognition Challenge}
\newacronym{gpu}{GPU}{graphics processing unit}
\newacronym{http}{HTTP}{Hyper Text Transfer Protocol}
\newacronym{osi}{OSI model}{Open Systems Interconnection model}
\newacronym{dns}{DNS}{Domain Name System}
\newacronym{html}{HTML}{Hypertext Markup Language}
\newacronym{api}{API}{Application Programming Interface}
\newacronym{url}{URL}{Uniform Resource Locator}
\newacronym{uri}{URI}{Uniform Resource Identifier}
\newacronym{json}{JSON}{JavaScript Object Notation}
\newacronym{xml}{XML}{Extensible Markup Language}
\newacronym{dna}{DNA}{Deoxyribonucleic acid}
\newacronym{nlp}{NLP}{natural language processing}
\newacronym{daisy}{DAISY}{Digital Automated Identification SYstem}
\newacronym{uml}{UML}{Unified Modeling Language}
\newacronym{db}{DB}{database}
\newacronym{hdd}{HDD}{hard disk drive}
\newacronym{ssd}{SSD}{solid state drive}
\newacronym{cpu}{CPU}{central processing unit}
\newacronym{ram}{RAM}{random access memory}
\newacronym{gif}{GIF}{Graphics Interchange Format}
\newacronym{gbif}{GBIF}{Global Biodiversity Information Facility}
\newacronym{os}{OS}{operating system}
\newacronym{jpeg}{JPEG}{Joint Photographic Experts Group}
\newacronym{captcha}{CAPTCHA}{Completely Automated Public Turing test to tell Computers and Humans Apart}

%% file: definitionen/variablen.tex
\newcommand{\titleDocument}{Master Thesis}
\newcommand{\subjectDocument}{M.Sc. in computer science / "Complex Software Systems"}

%% file: extras/deckblatt.tex
\thispagestyle{empty}
\begin{figure}[t]
 \centering
 \includegraphics[width=0.6\textwidth]{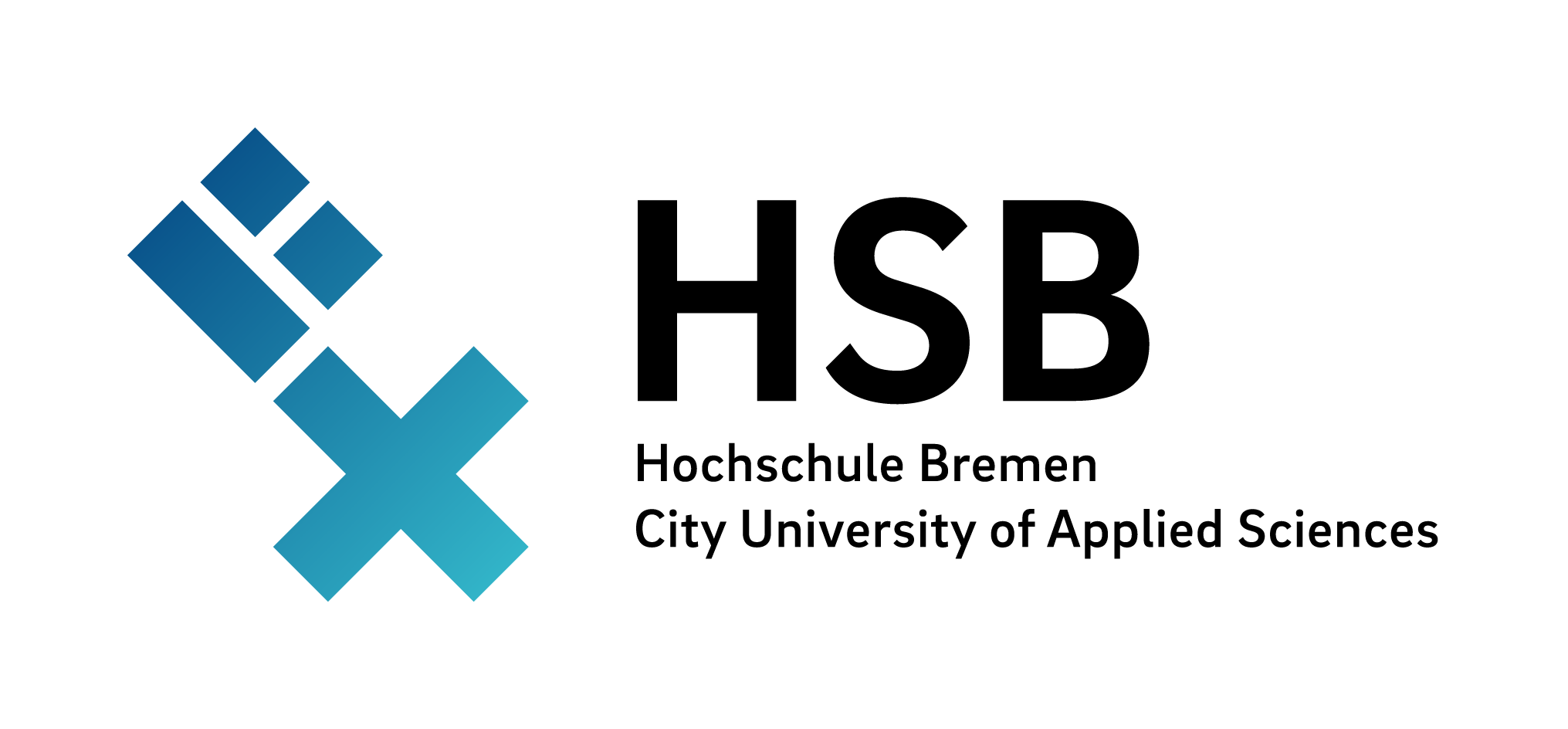}
\end{figure}
\begin{verbatim}

\end{verbatim}
\begin{center}
\Large{Hochschule Bremen}\\
\Large{- City University of Applied Sciences -}\\
\end{center}

\begin{center}
\Large{Faculty 4}
\end{center}
\begin{verbatim}

\end{verbatim}
\begin{center}
\doublespacing
\textbf{\LARGE{\titleDocument}}\\
\singlespacing
\begin{verbatim}

\end{verbatim}
\textbf{{~\subjectDocument~}}
\end{center}
\begin{verbatim}

\end{verbatim}
\begin{flushleft}
\begin{tabular}{llll}
\textbf{Topic:} & & Automated dataset generation for image & \\ 
& & recognition using the example of taxonomy & \\
& & \\
\textbf{Author:} & & Jaro Milan Zink & \\
& & Matrikel-No.: 295588 & \\
& & \\
\textbf{Version date:} & & 6th September 2017 &\\
& & \\
\textbf{Supervisor:} & & Prof. Dr.- Ing. Heide-Rose Vatterrott  &\\
\textbf{Co-supervisor:} & & Martin Winkler M.Sc. &\\
\end{tabular}
\end{flushleft}

%% file: extras/erklaerung.tex
\thispagestyle{empty}
\section*{Declaration of Authorship}\label{declaration_of_authorship}

I hereby declare that the thesis submitted is my own unaided work. All direct or indirect sources used are acknowledged as references.
\newline
\newline
I am aware that the thesis in digital form can be examined for the use of unauthorized aid and in order to determine whether the thesis as a whole or parts incorporated in it may be deemed as plagiarism. For the comparison of my work with existing sources I agree that it shall be entered in a database where it shall also remain after examination, to enable comparison with future theses submitted. Further rights of reproduction and usage, however, are not granted here.
\newline
\newline
This thesis was not previously presented to another examination board and has not been published.

\begin{displaymath}
\begin{array}{ll}
Signature:~~~~~~~~~~~~~~~~~~~~~~~~~~~~~~~~~~~~~~~~~~
& City, Date:~~~~~~~~~~~~~~~~~~~~~~~~~~~~~~~~~~~~~~~~~~
\end{array}
\end{displaymath}

%% file: extras/abstract.tex
\thispagestyle{empty}
\section*{Abstract}

This master thesis addresses the subject of automatically generating a dataset for image recognition, which takes a lot of time when being done manually. As the thesis was written with motivation from the context of the biodiversity workgroup at the City University of Applied Sciences Bremen, the classification of taxonomic entries was chosen as an exemplary use case. In order to automate the dataset creation, a prototype was conceptualized and implemented after working out knowledge basics and analyzing requirements for it. It makes use of an pre-trained abstract artificial intelligence which is able to sort out images that do not contain the desired content. Subsequent to the implementation and the automated dataset creation resulting from it, an evaluation was performed. Other, manually collected datasets were compared to the one the prototype produced in means of specifications and accuracy. The results were more than satisfactory and showed that automatically generating a dataset for image recognition is not only possible, but also might be a decent alternative to spending time and money in doing this task manually. At the very end of this work, an idea of how to use the principle of employing abstract artificial intelligences for step-by-step classification of deeper taxonomic layers in a productive system is presented and discussed.

%% file: kapitel/1_introduction.tex
\section{Introduction}\label{introduction}

This section will give an idea of the thesis' topic and explain why and how it was chosen. It will also describe the approached problem and furthermore trace out the aim of this dissertation.

\subsection{Problem description}

When working on a project involving image recognition or even \gls{ai_gls} in general, most tasks can be solved quite easily nowadays, due to a large number of available \glspl{framework} to be used. Some of them do not even require programming skills anymore and provide a graphical \gls{ui_gls}, where a scientist can just input some data and click a button to construct a fully functional \acrfull{ai}.

The most time consuming part of such a project is the creation of a dataset for the training phase. This task is hardly documented anywhere (in contrast to most other steps), which is probably because of the countless sources and data types available for different use cases; If one wanted to build an \acrshort{ai} being able to transform speech to text, the dataset would need to contain audio and text files, while someone working on image recognition needed pictures for the training.

In the field of \gls{ml_gls}, it can be stated in general terms that more data leads to a better result, but collecting and sorting thousands of images, for example, can be a lot of work if it is done manually and therefore can increase the cost factor of a project drastically.

For general object detection, there are already pre-sorted, manually gathered collections of images like ImageNet \cite{imagenet}, which even contain thousands of categories, but those collections usually follow a very broad approach of object classification and therefore can not be used in every scenario. To accomplish the targets of this thesis (see \ref{ssec:targets}), a more specialized dataset is required.

\subsection{Motivation / Background}

At the City University of Applied Sciences Bremen, there is a workgroup for \gls{biodiv}, consisting of people studying either biology or computer science. This workgroup has implemented many different projects in the past, one of those were the \textit{biodiversity atlases}. These atlases document (among other information) sightings of species made by the biology students and nature enthusiasts. 

But sometimes, a species cannot be determined easily, especially if one has not much experience in doing so. For example, there are many citizen scientists who want to contribute by reporting sightings, but often are not sure which species they just saw. To help identifying a species, the idea of using image recognition for this task came up relatively soon. The person would only need to take a picture of a given individual and the algorithm would give a result containing the lowest recognizable taxonomic layer, e.g. the family of the species.

Having a system which is able to do so would not only support new students in their process of learning, but also the many volunteers who participate in building and extending these openly available knowledge-pools. Currently, adding something to an atlas takes a lot of time, because one of the maintaining scientists needs to examine every new entry before it is published. If this step could be made less complex and lengthy by having a software preselecting and/or sorting out entirely wrong images, everyone involved in this project (and maybe even biologists around the world \cite{nature}) would benefit.

\subsection{Targets of this thesis} \label{ssec:targets}

The main goal is, as the title suggests, to solve the aforementioned problem by trying to automate the step of creating the dataset. To be able to do so, an exemplary use case is needed. As this thesis operates within the realm of biodiversity, the use case will involve taxonomic hierarchies to benefit the workgroup and may be used to develop a fully functional application.
\\
\\
There are two central questions to be answered by this publication:

\begin{itemize}
\item Is it possible to automate the creation of a dataset for training an \acrshort{ai} used for image recognition?
\item If yes: Which results does the automatically collected training-data give (in comparison to a manually created dataset)?
\end{itemize}

To be able to answer these questions, a concept for automatically acquiring a set of pictures (which, as mentioned above, is the training data when speaking about image recognition) shall be raised and implemented prototypically. When the implementation is completed, the generated dataset must be used to train an \acrshort{ai}, which can then be evaluated in order to find out if the principle of automating the dataset creation can be used in productive systems. The creation of the \acrshort{ai} requires a framework capable of doing so, which means that an appropriate one has to be found by comparing available candidates during the conceptional phase.

Other than the automated downloading of the pictures for the dataset, the most difficult challenge that will need to be overcome is the detection of unwanted images. When asking a search engine for pictures of ducks, for example, there will probably also be pictures of rubber ducks in the results. These must be filtered out so that there are only suitable images left in the automatically created dataset and the resulting \acrshort{ai} does not produce false positive recognitions.

\subsection{Structure}

This thesis is subdivided into eight chapters. In the first chapter, an overview of the work is given by presenting the statement of the problem, describing the motivation for the work and defining the targets of the thesis.

\vspace{6 mm}
The second chapter has the purpose to trace out the work's context. It illustrates the environment of the biodiversity warehouse the thesis is placed in and depicts similar publications from other researchers.

\vspace{6 mm}
The knowledge basics for the chosen topic are explained in the third chapter. This is done by giving insights into different fields and methods of \acrlong{ai}, disclosing the technique of web crawling and describing the science of taxonomy.

\vspace{6 mm}
In the fourth chapter, an analysis of the requirements for user, the prototypical system, the development hardware and the data(set) is done and discussed.

\vspace{6 mm}
A concept for the prototype is prepared within the fifth chapter, where tools for the different subareas of the application are chosen in consideration of the requirements defined before.

\vspace{6 mm}
The sixth chapters demonstrates the system setup, the prototypical implementation of the proposed system and the problems arising while doing so. Furthermore, code snippets are provided to give a detailed understanding of the software.

\vspace{6 mm}
The dataset that was automatically gathered by the prototype is evaluated in chapter seven. A concept for the evaluation is presented and a comparison against other, manually collected datasets is done.

\vspace{6 mm}
Eventually, a conclusion of the evaluation results is raised and future prospects for a productive system are offered in the eighth chapter.

%% file: kapitel/2_related_work.tex
\section{Related work}\label{related_work}

This passage shall outline the context of the chosen topic inside the biodiversity workgroup, give an overview of similar publications and set them in relation to this thesis.

\subsection{Biodiversity warehouse}

As mentioned before, the biodiversity workgroup is working with different systems. The biodiversity atlases are a part of a software conglomerate called the \textit{biodiversity warehouse}, which, as the name suggests, aims to provide as much information about biodiversity as possible. In the context of this warehouse many ideas, publications and implementations have arisen.  

The topics span a wide range and cover many areas of research. There is work regarding \acrfull{nlp} \cite{schrader}, descriptive data \cite{jeg}, gamification \cite{fritsche}, several other subjects regarding software development and automation with the help of \acrlong{ai} \cite{heil}. This thesis probably fits best inbetween the first and last of the aforementioned papers, as they also deal with a (different) form of \acrshort{ai} and try to find a solution for reducing the amount of manual work to be done by participators in the project.

\subsection{Other publications} \label{ssec:publications}

There are already systems like the \acrfull{daisy} \cite{daisy}, Pl@ntNet \cite{plantnet} or approaches like the one proposed by Alsmadi et al. \cite{fishrec}, being able to detect species by analyzing images and other data. But these systems rather focus on the productive detection and classification than building a dataset to do so.

Most publications about dataset creation for the training of an \acrshort{ai} involve human tasks, even when dealing with big data \cite{amzndata}. Google\footnote{\url{https://www.google.com/}} also admits to make use of their large human user base's ability to recognize text and objects by letting them solve \acrfullpl{captcha}:

\begin{quotation}
"\textbf{Powered by machine learning} \\
\textit{Hundreds of millions of \acrshortpl{captcha} are solved by people every day. reCAPTCHA makes positive use of this human effort by channeling the time spent solving \acrshortpl{captcha} into digitizing text, annotating images, building machine learning datasets. This in turn helps preserve books, improve maps, and solve hard AI problems.}"\footnote{\url{https://developers.google.com/recaptcha/}}
\end{quotation}

During the preparation to write this thesis, D. Grossman had a similar idea and wrote a blogpost about it \cite{grossman}. His approach deals with car models instead of taxonomic entries, but also utilizes an abstract \acrshort{ai}. It uses a more general model, which was already pre-trained using the ImageNet dataset in order to detect different objects. This means the abstract \acrshort{ai} cannot be used for tasks involving categories that are not included in the ImageNet. The approach of this thesis, in contrast, gives the possibility to create an \acrshort{ai} capabale to differentiate any kind of category. Also, there is no evaluation done in his publication where the proposed system is analysed whether it works correctly or not. This may be perfectly fine for writing a blogpost, but does not follow the guidelines of a scientific research.

%% file: kapitel/3_basics.tex
\section{Basics}\label{basics}

In this chapter, the thesis' emphases shall be discussed in detail. Therefore, the basics will be explained in particular on the following pages, so that a foundation of knowledge is given to understand and develop the next sections.

\subsection{Neural networks}

The origin of \acrfullpl{nn} are human beings themselves, because \acrshortpl{nn} are a mapping of brain cells in information technology. They're a digital representation of biological neurons, which are making up the majority of all known nerve-systems. Just like their biological counterparts, these artificial networks are capable of learning, which means they can produce output based on "experience" in form of a given input. To do this, \acrshortpl{nn} calculate the most probable output depending on the input data. As humans (or every other intelligent organism) follow the same pattern, \acrshortpl{nn} are considered an \acrlong{ai}.

In the context of computer science, \acrlongpl{nn} are especially helpful when looking at problems which do not have a clear structure. They also can process complicated calculations, but conventional algorithms are usually more efficient in doing so. Hence, they are mostly used to approximately give possible results, which can not easily be described by a simple formula.

\subsubsection{Architecture}

\begin{figure}[htbp]
    \centering
    \includegraphics[width=0.9\textwidth]{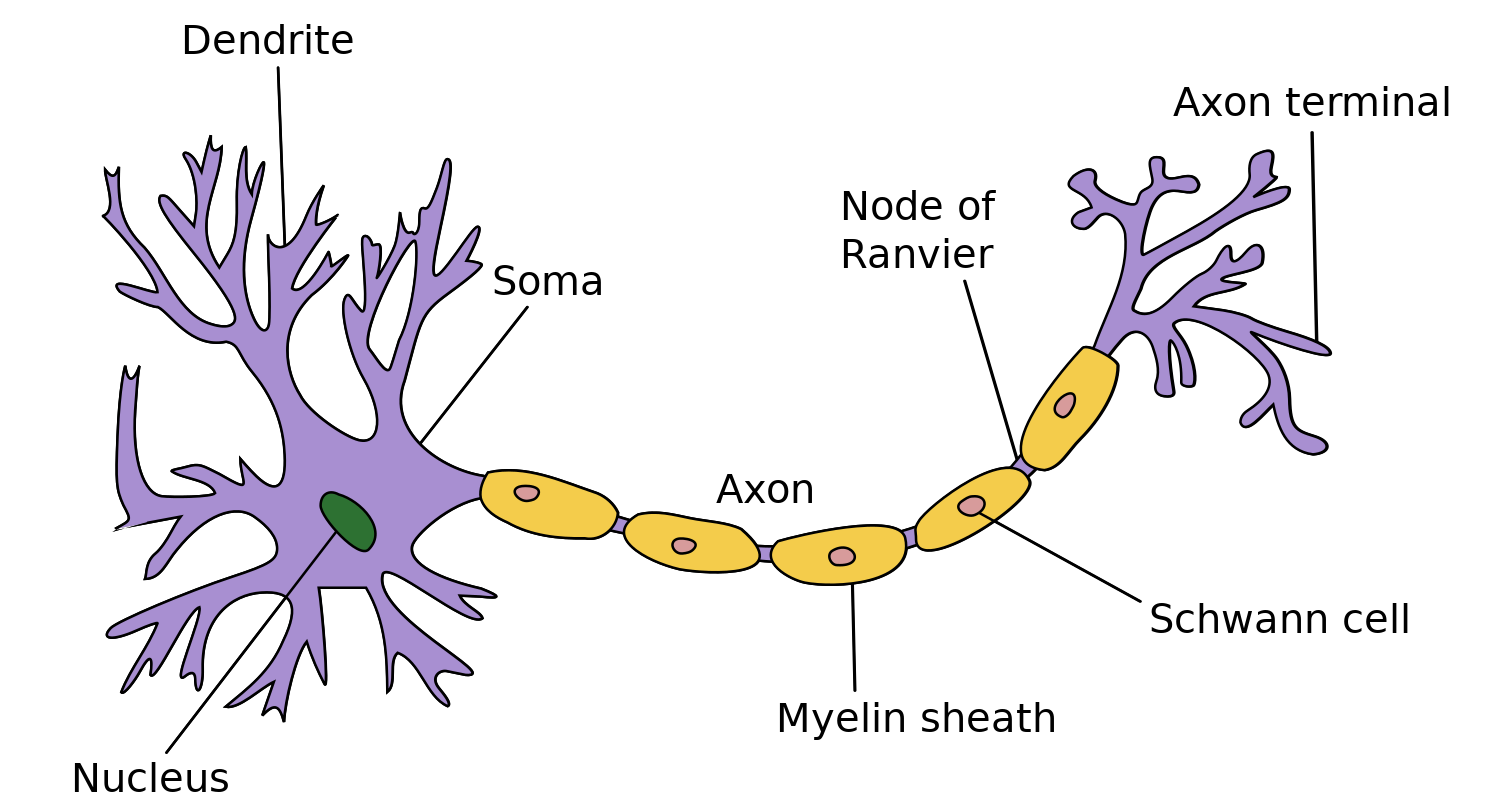}
    \caption[Structure of a typical neuron]{Structure of a typical neuron\footnotemark}
    \label{fig:bioneuron}
\end{figure}

Biological neurons are composed mainly of the Soma, Dendrites, Axon and Synapses (see figure~\ref{fig:bioneuron} [p. \pageref{fig:bioneuron}]). If a stimulus is created somewhere in the body, it is received by the Soma via the Dendrites, which will add up all incoming stimuli. If their sum excels a specific value, the Soma sends a signal to the neighbouring neurons over the Axon. The connections between two neurons are made of Synapses and are called the Axon terminal. The human brain exhibits a very high concentration of neurons in comparison to other mammals; it contains about 10\textsuperscript{11} of them altogether \cite[p.~133]{neuroscience}.

The sequence from a occurring stimulus to the neurons/brains reaction can be illustrated quite simply: Imagine a child touching a cooktop for the first time. When the hand comes in contact with the hot surface, its nerve-cells send many stimuli to the neurons inside the brain. If a given threshold value is reached, a chain reaction will be triggered from the neurons, which in turn initiates the natural reflex of contracting muscles (pulling the hand away) to avoid damage to the body. During this reaction, the child learns that a stove can be hot, because the neurons in the brain adjust the threshold and therefore it will pull the hand away sooner without touching the surface.

\begin{figure}[htbp]
    \centering
    \includegraphics[width=0.9\textwidth]{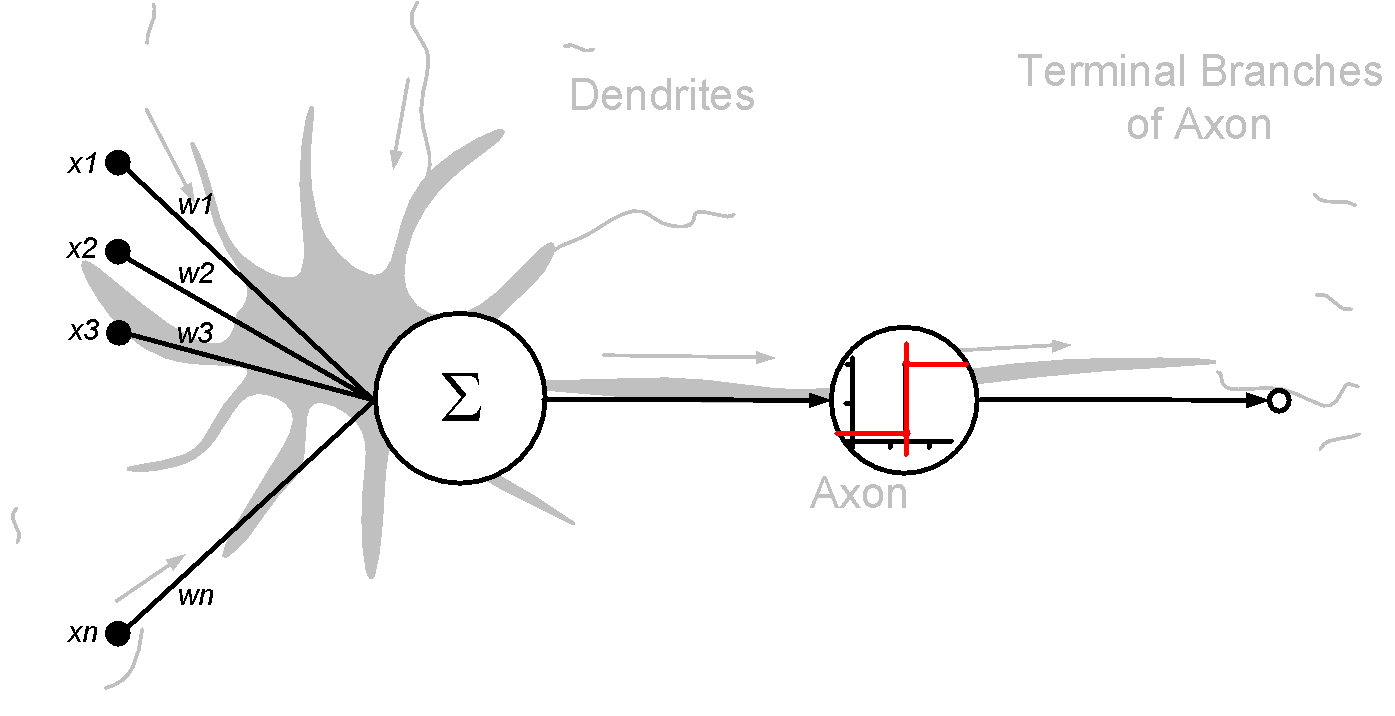}
    \caption[Structure of an artificial neuron]{Structure of an artificial neuron\footnotemark}
    \label{fig:aineuron}
\end{figure}

Generally speaking, artificial neurons are "built" almost the same as biological ones. As figure \ref{fig:aineuron} shows, they also have inputs (Dendrites), one module to add the inputs together (Soma), another module specifying a threshold (Axon) and, finally, an output (Axon terminal).

\footnotetext{\url{https://upload.wikimedia.org/wikipedia/commons/b/bc/Neuron_Hand-tuned.svg}}

\footnotetext{\url{http://www.nelsonrobotics.org/presentation_archive_nelson/nelson-intro-ann.ppt}, slide 8}

\newpage
The activation function of a neuron is monotonically increasing. A few examples for such a function are:

\begin{description}
\item[Hard-limit function]\hfill \\
Only can have the value of either 0 or 1 and is activated by any input $\geq 0$.
\item[Piecewise-linear function]\hfill \\
As the name suggests, this type of function is linear inside a specified interval and increases in respect to its input values until the maximum is reached. 
\item[Sigmoid function]\hfill \\
This function is used very often in \acrshort{nn}-implementations. It has a variable slope and is differentiable.
\end{description}

\begin{figure}[htbp]
    \centering
    \includegraphics[width=0.325\textwidth]{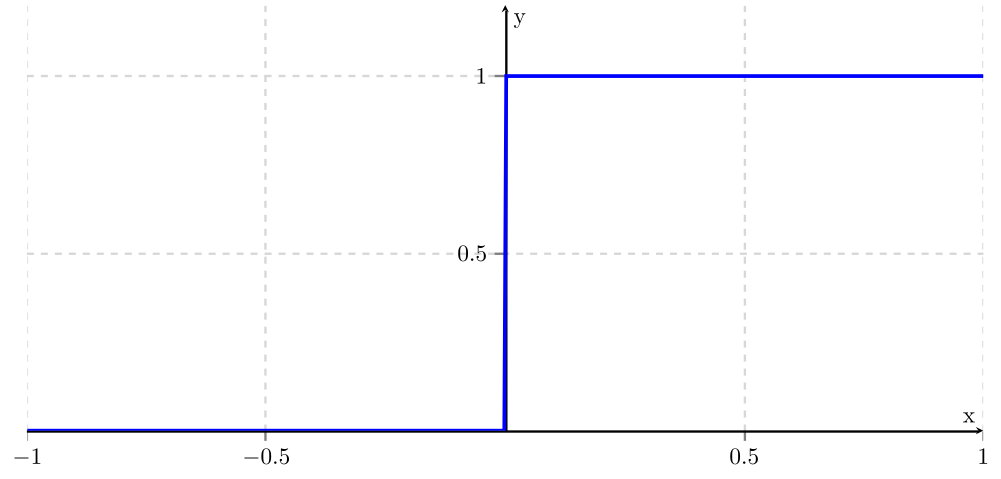}
    \includegraphics[width=0.325\textwidth]{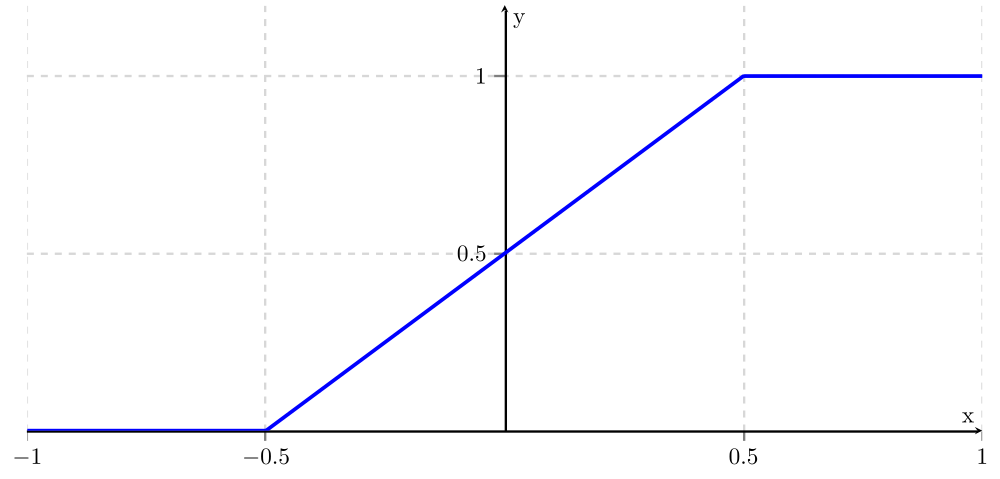}
    \includegraphics[width=0.325\textwidth]{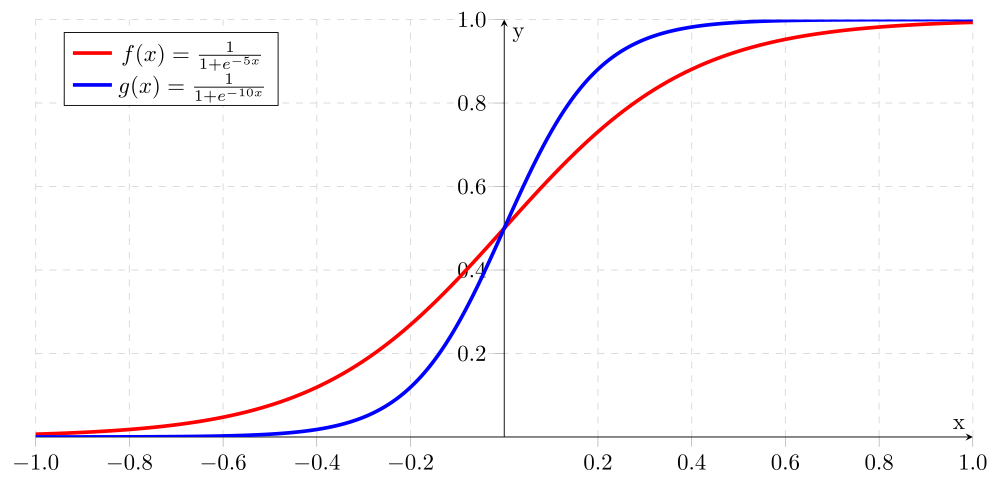}
    \caption[Activation functions]{Hard-limit\footnotemark, Piecewise-linear\footnotemark and Sigmoid\footnotemark -function (from left to right)}
    \label{fig:hlfunc}
\end{figure}

\addtocounter{footnote}{-3} 
\stepcounter{footnote}\footnotetext{\url{https://upload.wikimedia.org/wikipedia/commons/0/07/Hard-limit-function.svg}}
\stepcounter{footnote}\footnotetext{\url{https://upload.wikimedia.org/wikipedia/commons/6/6b/Piecewise-linear-function.svg}}
\stepcounter{footnote}\footnotetext{\url{https://upload.wikimedia.org/wikipedia/commons/f/f1/Sigmoid-function.svg}}

\begin{figure}[htbp]
    \centering
    \includegraphics[width=0.75\textwidth]{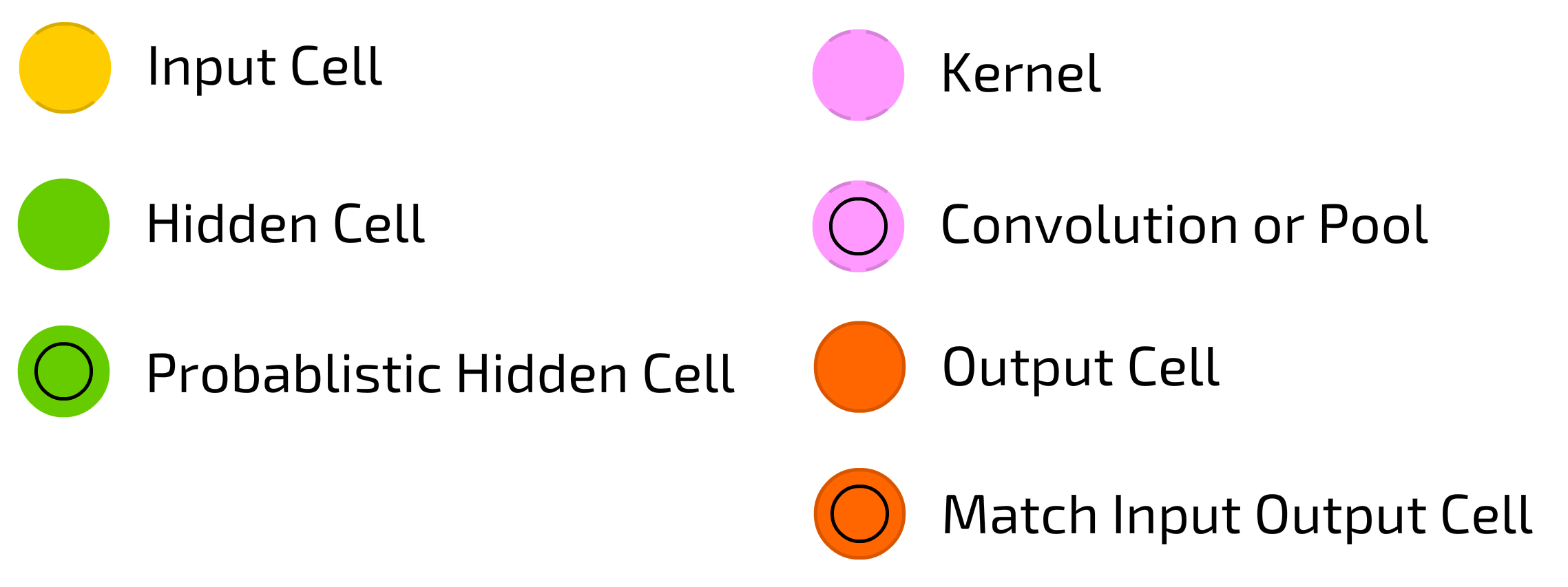}
    \caption[Neural network color legend]{Neural network color legend}
    \label{fig:nnlegend}
\end{figure}

\begin{figure}[htbp]
    \centering
    \includegraphics[width=0.425\textwidth]{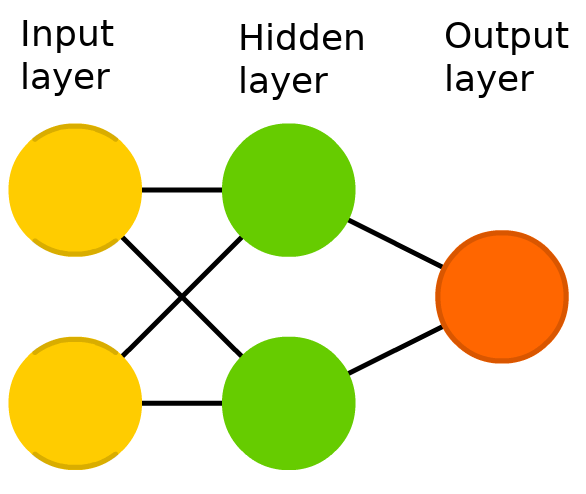}
    \caption[Layers of an artificial neural network]{Layers of an artificial neural network}
    \label{fig:nnlayers}
\end{figure}

In most cases, a neuron is not used on its own, but in a connected network, thus the name \textit{(artificial) \acrlong{nn}}. Such a network is usually subdivided into three layers; Incoming data is registered in the input-layer, which would be the nerve-cells inside a finger of the hand in the example above. At the end, there is an output-layer responsible for sending signals in form of data. This is represented by the reaction of pulling the hand away. In between those two layers, there are an unspecified number of layers for adding up the input, setting a threshold and calculating the output. These layers are referred to as \textit{hidden} layers. This architecture is shown in figure \ref{fig:nnlayers}, while figure \ref{fig:nnlegend} contains an explanation for the colors used to visualize \acrlongpl{nn} in the figures of this thesis.

\newpage
\subsubsection{Functionality}

Artificial \acrlongpl{nn} normally work in two phases. The first phase is called \textit{training phase} and describes the time period in which the network is "learning" while being fed with data given into the input layer. In the next phase, the \acrshort{nn} can already be used for its intended purpose. It will then calculate output using its own "knowledge".

Three major learning paradigms have been developed to train a \acrlong{nn}. They are briefly described in this list:

\begin{description}
\item[Supervised learning]\hfill \\
The most common method of supervised learning is \textit{backpropagation}. This means that the \acrlong{nn} gets a set of input- and desired output-data. Afterwards, the given, desired output data is compared to the one the \acrshort{nn} calculates itself. It will then learn by propagating back the rate of errors produced by this step to the input layer in order to adjust the weights of the connections between the single neurons. These weights are significant for the outcome of the net's further calculations.
\item[Unsupervised learning]\hfill \\
With this technique, the network does not get any predefined output values. It has to use the input data to categorize its neurons. It does so by activating random ones and finding the best fitting one. This neuron's (and its neighbour's) weights are then adjusted and the step is repeated. This process will form groups after a number of iterations depending on the size of the \acrshort{nn}. Unsupervised learning is usually implemented using so-called \textit{self-organizing maps} \cite{som}. 
\item[Reinforcement learning]\hfill \\
The \acrlong{nn} is fed with input data and will receive feedback whether its calculated outputs are right or wrong. It can use this information to adjust its weights during the training phase. This method is slower than supervised learning but provides a more general learning approach, which is often used to create \acrshort{ai} in video-games, for example.
\end{description}

\subsubsection{Implementations}

While there are many complex and advanced implementations of \acrshortpl{nn}, such as the Fast Artificial Neural Network Library\footnote{\url{http://leenissen.dk/fann/wp/}}, e.g., this subsection shall give an idea of a minimal one. The task of the network in this fictitious situation is to predict the output of a given input-vector (see table \ref{table:dataset} [p. \pageref{table:dataset}]). 

\begin{table}[htbp]
\begin{center}
\begin{tabular}{ |c|c|c|c| }
    \hline
    \rowcolor[gray]{.9} \multicolumn{3}{ |c| }{Inputs} & Output \\ \hline
    0 & 0 & 1 & 0 \\ \hline
    1 & 1 & 1 & 1 \\ \hline
    1 & 0 & 1 & 1 \\ \hline
    0 & 1 & 1 & 0 \\
    \hline
\end{tabular}
\caption{Example dataset}
\label{table:dataset}
\end{center}
\end{table}

The snippet below (Listing \ref{lst:minimalnn} [p. \pageref{lst:minimalnn}]) shows the whole source code needed to implement the \acrshort{nn} for this use case. Variable \textit{X} is assigned with the input dataset shown in the table before and can be thought of as the input layer. \textit{y} corresponds to the desired output and does not serve as its own layer. It is only used to calculate the error rate. \textit{l1} and \textit{l2} define the second (hidden) and third (output) layer of the network. The connections between these layers' neurons are represented by \textit{syn0} and \textit{syn1}, which contain the values of the weights.

\begin{minipage}{\linewidth}
\lstset{language=python, aboveskip=2.5em, belowskip=1.5em}
\begin{lstlisting}[frame=htrbl, caption={A very minimal implementation of a \acrfull{nn}}, label={lst:minimalnn}]
import numpy as np
X = np.array([ [0,0,1],[0,1,1],[1,0,1],[1,1,1] ])
y = np.array([[0,1,1,0]]).T
syn0 = 2*np.random.random((3,4)) - 1
syn1 = 2*np.random.random((4,1)) - 1
for j in range(60000):
    l1 = 1/(1+np.exp(-(np.dot(X,syn0))))
    l2 = 1/(1+np.exp(-(np.dot(l1,syn1))))
    l2_delta = (y - l2)*(l2*(1-l2))
    l1_delta = l2_delta.dot(syn1.T) * (l1 * (1-l1))
    syn1 += l1.T.dot(l2_delta)
    syn0 += X.T.dot(l1_delta)
print(l2)
\end{lstlisting}
\end{minipage}

In the \textit{for}-loop, the network will be trained using backpropagation. In each of the 60000 steps, an output is given by the \acrshort{nn}. Using that output, the weights are updated (line 11 \& 12) with the error rate calculated before (line 9 \& 10). This is a very clean and straightforward example for a backward propagation of errors. Afterwards, the call of the  \textit{print}-function with \textit{l2} as an argument produces the following output:

\lstset{language=python, aboveskip=2em, belowskip=1em}
\begin{lstlisting}[frame=htrbl, caption={Output of the minimal \acrshort{nn}}, label={lst:minimalnn_out}]
[[ 0.00213508]
 [ 0.99649038]
 [ 0.99508986]
 [ 0.00501845]]
\end{lstlisting}

If these numbers are rounded to the nearest whole number, the output from table \ref{table:dataset} is matched. It is thereby consequently shown that the \acrshort{nn} can forecast the desired values quite well. 

This simple illustration of a \acrlong{nn} was implemented and published by Andrew Trask \cite{trask}. However, the source code was slightly edited to match the Python3 syntax.

\subsubsection{Different types}

\begin{figure}[htbp]
    \centering
    \includegraphics[width=0.75\textwidth]{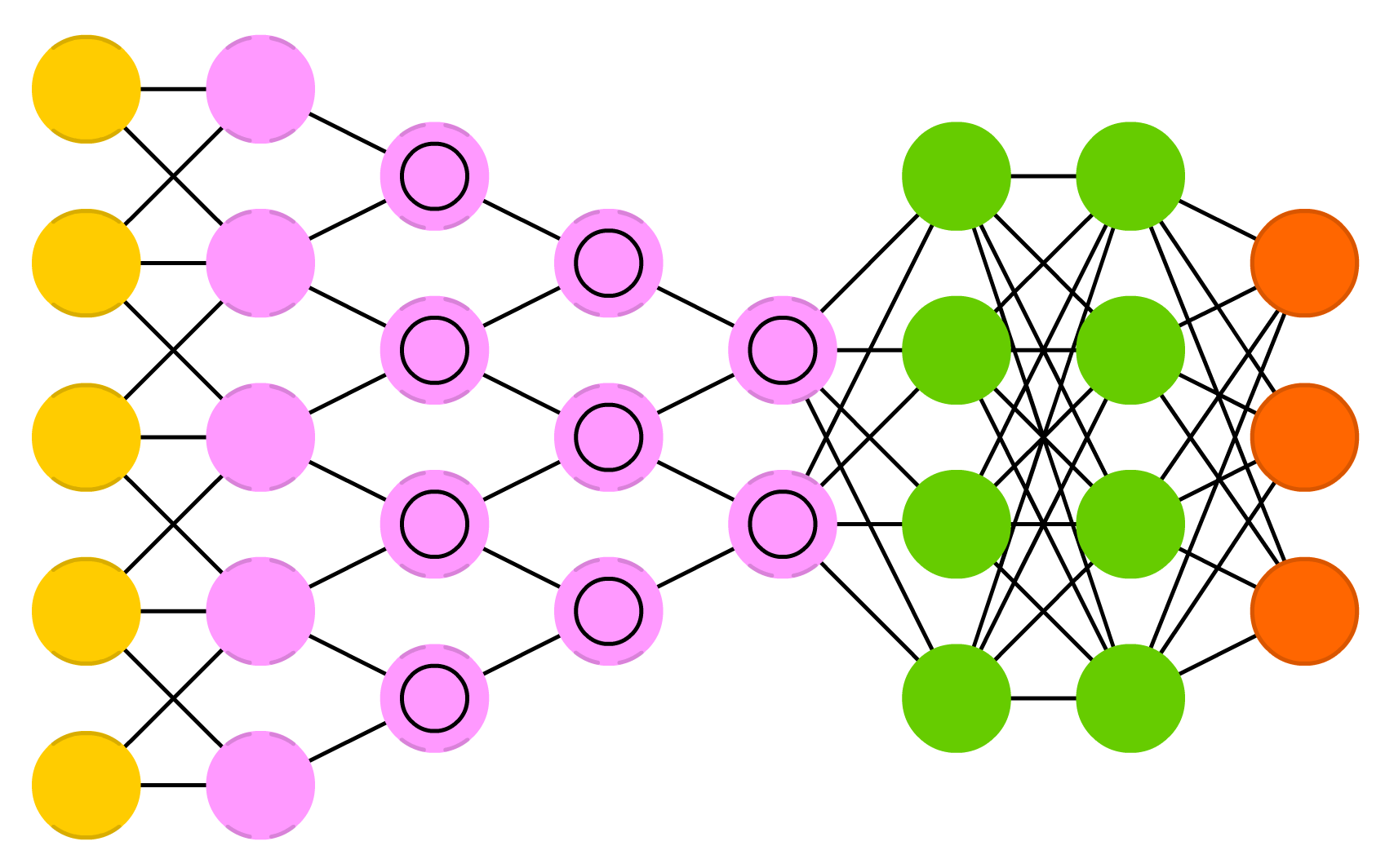}
    \caption[Sample of a (deep) convolutional \acrlong{nn}]{Sample of a (deep) convolutional \acrlong{nn}}
    \label{fig:dcnn}
\end{figure}

There are many different types of \acrlongpl{nn}. The subsections above describe a very basic one, which is purposed for trivial tasks and actions, as mentioned before. But when it comes to more sophisticated problems, more extensive structures are needed.
\newline
\newline
As this thesis will examine image recognition in particular, a \acrshort{nn} providing such a functionality may be a good example to begin with. Image recognition in terms of \acrlong{ai} describes the challenge of categorizing images by labelling them (see \ref{sssec:imgrec} for more details). For this task, the input data (pixels) is given into the network in overlapping segments. That means not each pixel of a picture is given in individually, but rather a square consisting of a predefined number of pixels. While the size of the square stays the same, the location of it is shifted by a few pixels for each input. With this technique, the whole image can be scanned and be fed to the input layer. Thereafter, the collected data is passed through (multiple) convolutional layers, where each neuron is only connected to its close neighbours. The next convolutional layer usually has fewer neurons than the one before. Also, these so-called \textbf{(deep) convolutional \acrlongpl{nn}} (figure \ref{fig:dcnn}) usually include one or more layers to pool similar pixels together. These \acrshortpl{nn} can also be trained to classify audio samples or other similar input data.

\begin{figure}[htbp]
\begin{minipage}{.4\textwidth}
  \centering
  \includegraphics[width=1\linewidth]{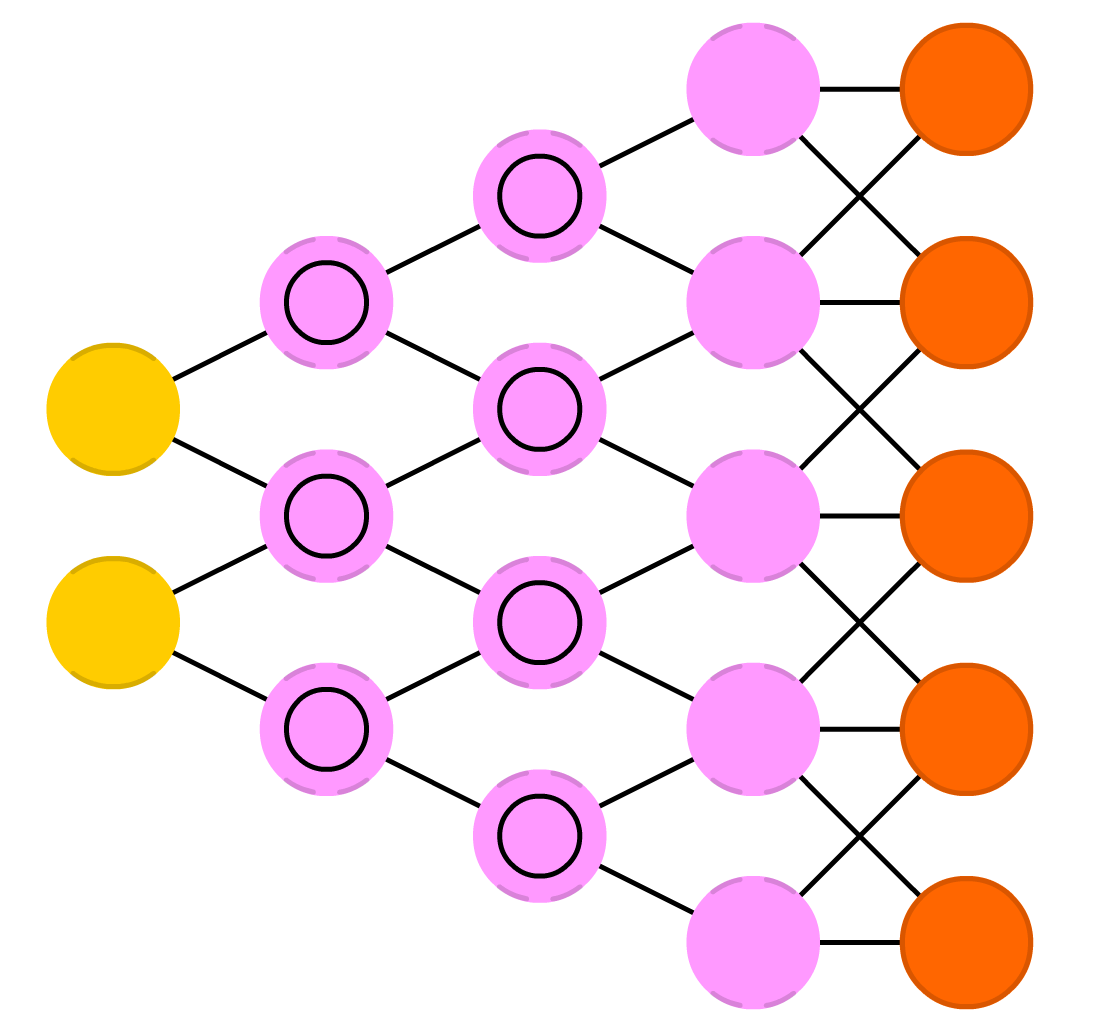}
\end{minipage}\qquad
\begin{minipage}{.55\textwidth}
  \centering
  \includegraphics[width=1\linewidth]{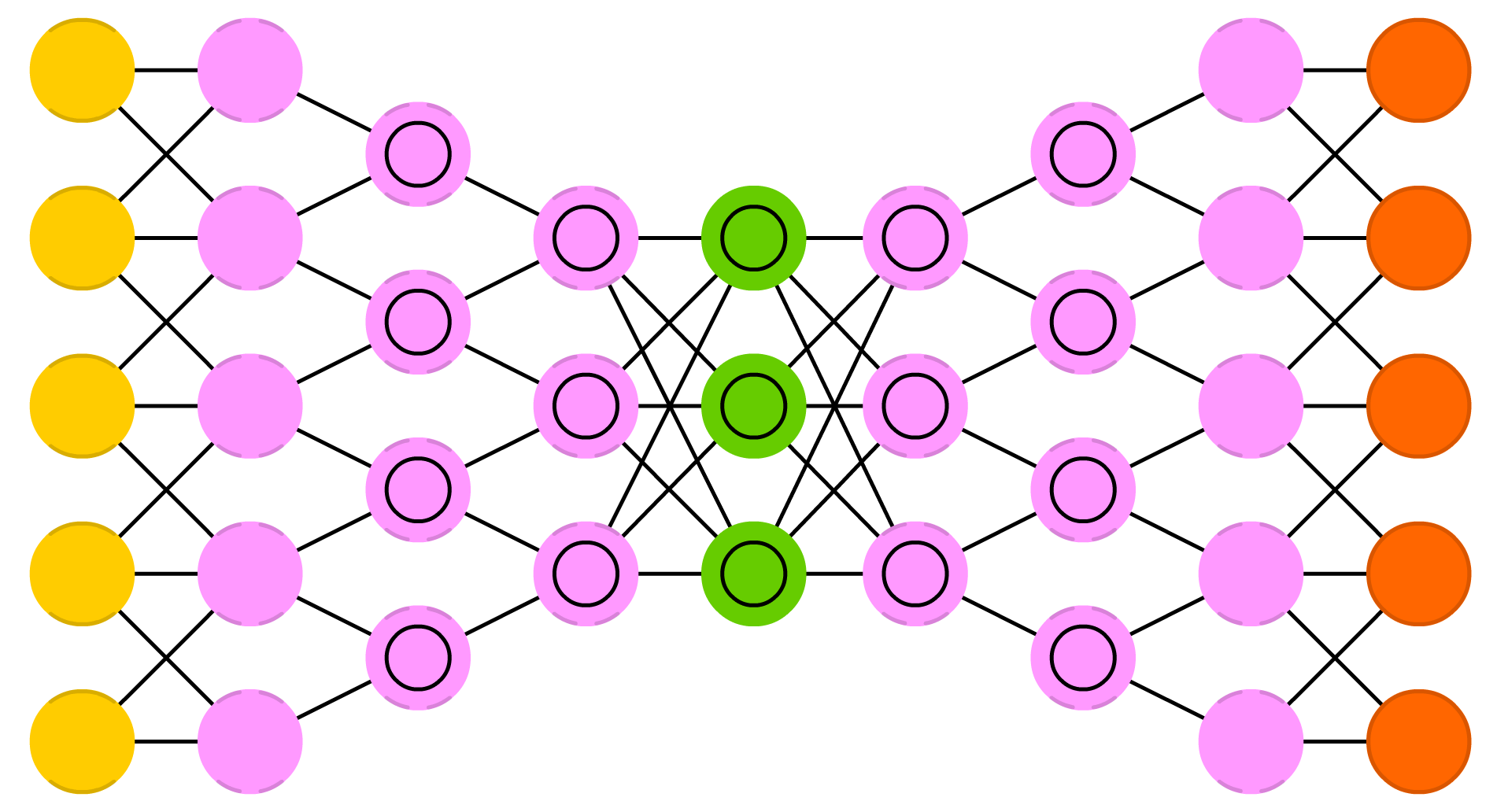}
\end{minipage}

\bigskip

\begin{minipage}[t]{.4\textwidth}
\centering
\captionof{figure}{Deconvolutional \acrlong{nn}}
\label{fig:dnn}
\end{minipage}\qquad
\begin{minipage}[t]{.55\textwidth}
\centering
\captionof{figure}{Deep convolutional inverse graphics network}
\label{fig:dcign}
\end{minipage}
\end{figure}

This task can also be reversed. For example, \textbf{deconvolutional \acrlongpl{nn}} (figure \ref{fig:dnn}) can be trained to produce images based on words. If one would want to take this one step further and combine both of those types, a \textbf{deep convolutional inverse graphics networks} (figure \ref{fig:dcign}) would be created. These \acrshort{nn} are capable of manipulating images in terms of removing / adding trained objects, or even rotating 3D objects, as proven by \cite{dcign}. 

\newpage
There are many more types of \acrlongpl{nn}. This subsection was only to give brief idea of how \acrlongpl{nn} can be modified for particular needs. The forestanding figures were taken off this graphic\footnote{\url{http://www.asimovinstitute.org/wp-content/uploads/2016/09/networkZooPoster.png}}. It is part of a blogpost describing and visualizing most currently existing \acrshort{nn}-types \cite{veen}.

\subsection{Machine learning}

The term \acrfull{ml} stands for the whole process of creating an \acrlong{ai} by collecting data, writing code and training a \acrlong{nn}. Often, the \acrshort{ai} has to be optimized and/or re-trained during several iterations until it is able to serve its purpose efficiently. The aforementioned steps sometimes can be automated though, so that a fully autonomous \acrlong{ml} system only needing a dataset as an input is produced.

\Acrlong{ml} is furthermore often used as a general term to sum up different subareas of itself. For instance, articles with content on speech recognition are often tagged with \acrshort{ml} to reach a broader audience. The following subsections shall give information about a few of these subareas which are essential for this thesis.

\subsubsection{Computer vision}

Computer vision expresses the ultimate goal of enabling devices with a human-like vision. That means a computer shall be able to capture its surroundings through sensors, process and set the values gathered by them in relation to each other and finally understand and react to the situation it measured. Computer vision is not really a subarea of \acrlong{ml}, but is strongly connected to it. The process of CV usually involves some form of \acrshort{ml}, such as image processing/recognition, speech to text conversion, etc. In addition, it employs the use of other algorithms to calculate the aforementioned relations, like the distance between itself and a (previously recognized) car. Afterwards, it can react by slowing down or warning people, depending on its purpose being autonomic driving or helping to park, e.g.

\subsubsection{Image processing and recognition} \label{sssec:imgrec}

Image processing and image recognition both describe the same subarea of \acrlong{ml}. It could be argued that image processing stands for the manipulation of the picture data, while image recognition aims to detect the images' content. Processing does not necessarily involve \acrshort{ml}, because many trivial tasks can be solved by specifically adjusted algorithms (such as taking the color away from a photo to make it grey scaled for example) and recognition describes more complex tasks (like image classification, pattern recognition and face-/object-detection), which always involve some kind of \acrlong{ai}. But since there is no clear definition to divide the two terms easily and it seems to fit the purpose better, the term \textit{image recognition} will be used in this work.

\subsubsection{Deep learning} \label{sssec:deeplearning}

Deep learning outlines the technique to create larger \acrlongpl{nn} by adding more layers and neurons, while also increasing the amount of training data. With this method, \acrlongpl{ai} have become much more powerful and human-like, so that today it is used in almost every \acrlong{ml} application. It was perhaps primarily practically realized in 2012, when a huge network containing 650000 neurons was proposed and tested on a \acrfull{gpu} for the first time \cite{deep1}. This approach was adopted and improved later that year to achieve an top-5 error rate of 15,8\% in the \acrfullpl{ilsvrc}\footnote{\url{http://www.image-net.org/challenges/LSVRC/}} by being run through a network with 1 billion connection on a cluster of 1000 machines and trained on 10 million images \cite{deep2}. The term \textit{top-X error} describes a result given by a \acrshort{nn} where all of the demanded X predicted categories are wrong. For example: If a picture contains only a tomato, but the five predictions are apple, peach, potato, onion and pear, the result represents a top-5 error. In most tests, X is set to five, but it can also be any other number that is lower than or equals the amount of categories the \acrshort{nn} was trained on. The top-X error \textit{rate} is the overall percentage of errors on a given dataset.

\subsubsection{Fields of application}

\Acrlong{ml} is used in many different areas. In fact, there are so many possible applications that they probably can not all be described and explained in this chapter. Instead, some examples will be given in the following enumeration.

\begin{description}
\item[Healthcare]\hfill \\
In the medical sector, image recognition is used to detect and diagnose different diseases, such as cancer \cite{health1} or diabetes \cite{health2}. The \acrshort{ai} can be either being brought in directly into the doctor's office and assist right there whilst doing a screening or make assumptions based on pictures send to it remotely. \newline
Another field where \acrshort{ml} is used for healthcare is drug discovery. The \acrlong{ai} can assist in finding the right medicine and dose or confirm the doctor's choice by evaluating many more previous cases and their outcome than any human could do in their entire lifetime \cite{health3}. 
\item[Military]\hfill \\
Of course, there are non-civilian \acrshortpl{ai}, too. The most prominent ones are presumably drones, or \textit{unmanned aerial vehicles (UAVs)}, as they are called officially. Drones assumedly make use of different \acrlong{ml} techniques, but the most well-known is image processing. They use it to patch huge amounts of pixels together to one large surveillance picture \cite{mil1} or to track vehicles in wide areas \cite{mil2}. There are also rumors that newer models are able to detect faces or at least track different persons \cite{mil3}, but these have not been confirmed yet at the time of writing. 
\item[Financial sector]\hfill \\
Most trades on stock markets have been made by some form of algorithm or \acrshort{ai} for years. Machines can predict stock prices \cite{fin} and decide to buy or sell them much faster than human brokers. With more and more companies using these technologies, soon a high-frequency trading market evolved. This again led to a contest in having the lowest response times, so that there even was built a new fiber cable connection between Chicago and New York \cite{fin2}, just because of \acrlong{ai}.\\
\\
\item[Virtual assistants]\hfill \\
Virtual assistants are probably the most common, yet unnoticed form of \acrshort{ai}. Millions of people use them every day without knowing that it is a \acrlong{nn} giving them their requested information about the weather, nearest cinema or upcoming appointments. The software called Siri\footnote{\url{https://www.apple.com/ios/siri/}}, Google Now\footnote{\url{https://www.google.com/intl/en-GB/landing/now/}}, Cortana\footnote{\url{https://www.microsoft.com/windows/cortana}} or Alexa\footnote{\url{https://developer.amazon.com/alexa}} aims to give an alternative to interact naturally with a device and thus minimizing the learning curve. The user shall have the impression of talking to another person when using it, so that it is possible to do other things in the meantime.
\item[Art]\hfill \\
As addressed before, one example for the use of \acrlong{ml} in art would be a style transfer, where an image receives the stylistic appearance of another one \cite{art}. The colorization of old back and white photographs is a specific use case where an \acrlong{ai} can help a human professional to reduce the amount of work and time \cite{art2}. Another example would be the removal of watermarks \cite{watermarks}. But pictures aren't the only form of art there is, of course. With \acrshort{ml}, creating music and sound has also been simplified, as two experiments\footnote{\url{https://aiexperiments.withgoogle.com/drum-machine}}\textsuperscript{,}\footnote{\url{https://aiexperiments.withgoogle.com/ai-duet}} have shown, e.g.
\item[Information security]\hfill \\
In 2016, there was a hacking-tournament where only \acrshortpl{ai} could participate\footnote{\url{http://archive.darpa.mil/cybergrandchallenge/}}. The machines had to secure themselves against attacks from the other participants. All of them had been given specific bugs so that there was a security vulnerability needed to be closed. But there was also a bug unknown to the hosts of the event discovered and used by one machine to attack another. A third noticed this and reacted by reverse engineering the attack, fixing the bug in a patch and applying this to itself. (translated from \cite{sec})
\end{description}

In the current year (2017), there even was a \acrshort{nn} created with the purpose to optimize other \acrlongpl{nn} called \textit{Google AutoML} \cite{mlnn}. It was able to solve this task better than its creators from Google, who are seen as experts and pioneers in this industry. This seems to be another step towards singularity, which defines the point in time where machine will supersede humans in being the highest form of intelligence on earth.

\subsection{Web crawling}

Gathering information from many (maybe different) websites is not done by hand anymore nowadays. Since Larry Page had written a first version of the Googlebot\footnote{\url{https://groups.google.com/forum/\#!msg/comp.lang.java/aSPAJO05LIU/ushhUIQQ-ogJ}} in 1996, which probably requested and indexed more websites than any other, crawlers have become an integral part of the internet. Per definition, a crawler is "\textit{A program that systematically browses the World Wide Web in order to create an index of data}"\footnote{\url{https://en.oxforddictionaries.com/definition/crawler}}. Mostly used in search engines, they can also be implemented to do many other things, such as downloading files, collecting information on a specific topic or checking for a website or other specific information to become available in a preset interval, for example. 

\begin{figure}[htbp]
    \centering
    \includegraphics[width=1.0\textwidth]{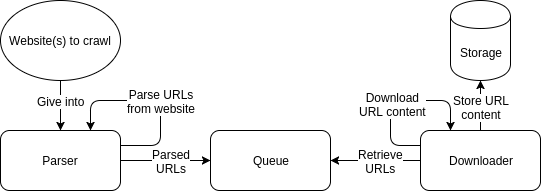}
    \caption[Schematic representation a crawler's architecture]{Schematic representation a crawler's architecture}
    \label{fig:crawler}
\end{figure}

A crawler is usually given one or more websites to work with. It parses the website's source code and extracts the \acrfullpl{url} of different elements, for example images or videos, depending on its purpose. The resulting \acrshortpl{url} are put into a queue, which will be polled by a downloader. Every time the downloader retrieves and deletes an item from the queue, the \acrshort{url}'s content will be downloaded and stored persistently.

Figure \ref{fig:crawler} shows a schematic representation of the architecture of a simple crawler, but to fully understand how they work, it is important to look at the underlying techniques being applied by them in detail. This shall be pointed out in the following subsections. (Note: The 7-Layer \acrfull{osi}\footnote{\url{https://en.wikipedia.org/wiki/OSI_model}} and \acrfull{dns}\footnote{\url{https://en.wikipedia.org/wiki/Domain_Name_System}} will not be explained, because that would make this section unnecessarily complex and go beyond the scope of this thesis)

\subsubsection{HTTP requests}

In order to get the contents of a website, every device uses the standardized \acrfull{http} to send requests to the \acrshort{http}-server hosting the site. A standard \acrshort{http} request sent through curl\footnote{\url{https://curl.haxx.se/}} is shown in listing \ref{lst:req1}, where lines preceded by * are handling the connection, those beginning with > contain the actual request and lines starting with < stand for the response given by the server.

The most important and only mandatory lines are the first two of the request. GET defines the request method, / is the requested path/resource on the host called \textit{\acrfull{uri}} and HTTP/1.1 determines the protocol version. Host: hs-bremen.de sets the (virtual) hostname to be resolved on the server, which is needed since \acrshort{http} version 1.1.

\begin{minipage}{\linewidth}
\lstset{language=bash, aboveskip=2em, belowskip=1em}
\begin{lstlisting}[frame=htrbl, caption={Output of a sample request to hs-bremen.de, using curl}, label={lst:req1}]
* Rebuilt URL to: hs-bremen.de/
*   Trying 194.94.24.5...
* TCP_NODELAY set
* Connected to hs-bremen.de (194.94.24.5) port 80 (#0)

> GET / HTTP/1.1
> Host: hs-bremen.de
> User-Agent: curl/7.54.1
> Accept: */*

< HTTP/1.1 301 Moved Permanently
< Date: Sun, 18 Jun 2017 15:00:41 GMT
< Server: Apache
< Location: http://www.hs-bremen.de/internet/de/
< Content-Length: 306
< Content-Type: text/html; charset=iso-8859-1
\end{lstlisting}
\end{minipage}

In this case the server sent only back headers without content, because the response code 301 means that the requested resource has been moved somewhere else. In this case, modern browsers use the given location header to request the new path. Curl can also do this, but it doesn't until it is told to by using a specific parameter. 301 is neither bad nor good, but the best case when requesting is a status code with the value of 200, meaning everything is OK and will be served as requested. For the sake of completeness, however, the following table shall give an overview over the most common response codes in \acrshort{http}.

\begin{table}[htbp]
\begin{center}
\begin{tabular}{ |c|c| }
    \hline
    \rowcolor[gray]{.9} Status code & Describing phrase \\ \hline
    200 & OK \\ \hline
    301 & Moved permanently \\ \hline
    302 & Moved temporarily \\ \hline
    403 & Forbidden \\ \hline
    404 & Not found \\ \hline
    500 & Internal server error \\ \hline
    503 & Service unavailable \\ \hline
    504 & Gateway timeout \\ \hline
    
\end{tabular}
\caption{\acrlong{http} status codes}
\label{table:httpcodes}
\end{center}
\end{table}

Besides GET, there are several other request methods available for \acrshort{http} and each of them stands for a specific action to be performed. Table \ref{table:httpmethods} (p. \pageref{table:httpmethods}) lists each of these methods and their associated descriptions, as defined in their specifications.

\begin{table}[htbp]
\begin{center}
\begin{tabular}{ |p{3cm}|p{11cm}| }
    \hline
    \rowcolor[gray]{.9} Method & Description \\ \hline
    GET & Transfer a current representation of the target resource. \\ \hline
    HEAD & Same as GET, but only transfer the status line and header section. \\ \hline
    POST & Perform resource-specific processing on the request payload. \\ \hline
    PUT & Replace all current representations of the target resource with the request payload. \\ \hline
    DELETE & Remove all current representations of the target resource. \\ \hline
    CONNECT & Establish a tunnel to the server identified by the target resource. \\ \hline
    OPTIONS & Describe the communication options for the target resource. \\ \hline
    TRACE & Perform a message loop-back test along the path to the target resource. \\ \hline
    PATCH & Requests that a set of changes described in the request entity be applied to the resource. \\ \hline
\end{tabular}
\caption{\acrshort{http} request methods\protect\footnotemark\textsuperscript{,}\protect\footnotemark}
\label{table:httpmethods}
\end{center}
\end{table}

\footnotetext{\url{https://tools.ietf.org/html/rfc7231\#section-4}}
\footnotetext{\url{https://tools.ietf.org/html/rfc5789\#section-2}}

As web crawlers are mostly used to retrieve data, they normally only make use of the GET-method. That doesn't mean they can not use other methods like POST to log in/authenticate, for example, but such a case happens certainly infrequent. 

\subsubsection{HTML}

When a browser is ordered to display a website, the first thing it does is sending a \acrshort{http} GET request. The response body will then, if the status code is 200, contain text in a special format: the \acrfull{html}. Since it was first suggested by Tim Berners-Lee in 1989 \cite{html}, \acrshort{html} quickly gained attention and acceptance and soon became the standard way to define a website(-layout). This language is designed to describe different parts of a website in a way the browser can "understand" easily, minimizing the chance of errors.

The only required entry of a valid \acrshort{html} layout is (from version 5 onwards) the document type. In practice, a basic website is formatted as shown in listing \ref{lst:html} (p. \pageref{lst:html}). It then also contains \textit{html}-, \textit{head}-, \textit{title}- and \textit{body}-tags, so that the browser has something to display.

\begin{minipage}{\linewidth}
\lstset{language=html}
\begin{lstlisting}[frame=htrbl, caption={Basic \acrshort{html} website layout}, label={lst:html}]
<!DOCTYPE html>
<!-- A comment -->
<html>
  <head>
    <title>Test</title>
  </head>
  <body>
    Website content
  </body>
</html>
\end{lstlisting}
\end{minipage}

There are many more elements a \acrshort{html} layout can include, but the one most worth mentioning in the context of this thesis is the \textit{img}-tag. As the name suggests, it is used to make a browser display an image by linking to it inside a source-attribute. Based on this mechanism, a web crawler can be programmed to grab only the \textit{src}-attributes of all image tags inside a given website, in preparation to download all the linked images. And as this is a standard every website has to implement, the image-crawler will be able to process all of them.

\subsubsection{APIs}

While websites written in \acrshort{html} are mainly designed to fit the human eye, there is another form of web resource, built for the communication between machines only. If the operator of a web service wants to give a possibility for automated access to certain data, the service can provide a so-called \acrfull{api}. These interfaces are accessed (just like websites) through a \acrshort{http} request to a given \acrshort{url}. But instead of \acrshort{html} to be rendered and visualized, an \acrshort{api} returns data in text form to be interpreted and processed by the requesting program. In most cases, the response is formatted in either the \acrfull{json} or \acrfull{xml}, but the latter seems to be deprecated by many developers in favor of the more efficient \acrshort{json}, as it has less overhead data and therefore causes less traffic. Also, \acrshort{json} uses a more object-oriented approach of data description, so that it costs less computing resources to parse it in many use cases. Listing \ref{lst:api} shows the \acrshort{json}-response of an exemplary \acrshort{http} request to an \acrshort{api}.

\begin{minipage}{\linewidth}
\begin{lstlisting}[frame=htrbl, caption={Example result of an \acrshort{http} GET request to an \acrshort{api}\protect\footnotemark}, label={lst:api}]
{
    ip: "127.0.0.1"
}
\end{lstlisting}
\end{minipage}

\footnotetext{\url{https://api.ipify.org/?format=json}}

\Acrlongpl{api} are especially important to developers creating client software, because they can work with a entirely defined resource which will not suddenly change its behaviour. The source code of websites, in contrast, can be modified several times a week (without any changes to their appearance). Hence, programs parsing the \acrshort{html} could fail to do so, because one of the elements vanished, e.g. \acrshort{api} providers usually notify their users a long time before making breaking changes or they just release a new version which can be found under another \acrshort{uri}, so that developers have enough time to react.

\subsection{Taxonomy}

Generally speaking, the word \textit{taxonomy} only expresses "\textit{the science or practice of classification}"\footnote{\url{https://www.collinsdictionary.com/dictionary/english/taxonomy}}, which can be applied in many different scientific fields. Taxonomy uses the organizational scheme of a nested hierarchy, meaning that every level but the top one is encompassed in its parent. There can exist multiple items inside each level, but every item can only have one parent. The following equations shall depict this scheme in a mathematical way:

\[ 	Duck \subset Birds \subset Animals \subset Organisms \]
\[ 	Beaver \subset Mammals \subset Animals \subset Organisms \]

Birds and Mammals both have multiple children, which are not related to each other, but both have the same parent: Animals. Organisms, in turn, also include Animals and its siblings. But as Organisms are in the top layer of this hierarchy, it its the only one not contained in a parent.

The term \textit{taxonomy} originally described the classification of organisms in biology, which also is most suitable for this work and will be referred to as this from now onward. The word itself derives from the ancient Greek "taxis" ('arrangement', 'order') and "nomia" ('method') or "nomos" ('law', 'managing')\footnote{\url{http://www.biology-online.org/dictionary/Taxonomy}}. The first one to classify organisms was Aristotle, whose method was irreplaceable until in 1758, Carolus Linnaeus invented the \textit{Linnaean system} \cite{tax1}. This system introduced a standard to describe different levels in the taxonomic hierarchy and is still used today with a few modifications. Today's major taxonomic ranks are listed (in hierarchic order) in table \ref{table:taxonomy} (p. \pageref{table:taxonomy}). In zoology, there are also minor ranks, such as: \textit{subfamily, subgenus, subspecies}, but these are not part of the standard system.

\begin{table}[htbp]
\begin{center}
\begin{tabular}{ |c|c| }
    \hline
    \rowcolor[gray]{.9} Latin & English \\ \hline
    vitae & life \\ \hline
    regio & domain \\ \hline
    regnum & kingdom \\ \hline
    phylum & phylum \\ \hline
    classis & class \\ \hline
    ordo & order \\ \hline
    familia & family \\ \hline
    genus & genus \\ \hline
    species & species \\ \hline
\end{tabular}
\caption{Taxonomic ranks}
\label{table:taxonomy}
\end{center}
\end{table}

In taxonomy, items of a layer are divided into different \textit{taxa} (singular: \textit{taxon}) by looking at specific taxonomic attributes. These attributes have been only of morphologic nature, until Charles Darwin conceived the theory of evolution and it could be used to define relations between organisms of the same ancestry. In 1966, Willi Hennig described a technique called \textit{cladistic analysis}, which is also known as \textit{phylogenetic systematics} \cite{tax2} and has become a standard for the classification of organisms. The newest contribution to the pool of classification procedures is of course the analysis of \acrfull{dna}. Modern cladistic alogrithms use the information from a individual's \acrshort{dna} to calculate the difference in the genetic constitution of certain species and build a phylogenetic tree with the results \cite{tax4}. These trees can be seen as a visualization of taxonomic links.

\begin{table}[htbp]
\begin{center}
\begin{tabular}{ |c|c|c|c| }
    \hline
    \rowcolor[gray]{.9} Taxonomic rank & Example 1 & Example 2 & Example 3 \\ \hline
    domain & Eukaryota & Eukaryota & Eukaryota \\ \hline
    kingdom & Animalia & Animalia & Animalia \\ \hline
    phylum & Chordata & Chordata & Chordata \\ \hline
    class & Aves & Mammalia & Actinopterygii \\ \hline
    order & Anseriformes & Rodentia & Perciformes \\ \hline
    family & Anatidae & Castoridae & Gobiidae \\ \hline
    genus & Anas & Castor & Neogobius \\ \hline
    species & Anas platyrhynchos & Castor fiber & Neogobius melanostomus \\ \hline
\end{tabular}
\caption{Examples of taxonomic classifications}
\label{table:taxexample}
\end{center}
\end{table}

Each classified species is assigned to every taxonomic rank through its taxa, as shown with the example of the species \textit{Castor fiber} (European Beaver) in table \ref{table:taxexample}. Because the taxa are defined and differentiated by taxonomic attributes, species with more similarities also share more taxonomic ranks. For example, many insects have six legs, but are still very different in their appearance and behaviour. Nonetheless, they all belong to the taxon "Hexapoda" in the (zoologic) taxonomic rank "Subphylum". 

In general, it can be stated that with a higher taxonomic rank, the similarities decrease, so that the count of taxa in the rank "kingdom" is only 6 (in the proposed system of Thomas Cavalier-Smith \cite{tax3}), e.g.

%% file: kapitel/4_requirements_analysis.tex
\section{Requirements analysis}\label{requirements_analysis}

On the following pages, the requirements for a successful realization of the method of resolution proposed in the first chapter shall be analyzed. As the contemplated application is thought of as rather a proof-of-concept implementation than a fully functioning program, there are not as many requirements to analyze, though.

\subsection{User requirements}

The user, who equals the developer/researcher in this scenario, does not require more than a simple interface in form of a command line without a graphical \acrshort{ui} to be able to interact with the application. Contrary wise, there are some preparations to be done by the user in order to run the prototype, because there will be some not automatable steps needed to be done by hand.

\begin{figure}[htbp]
    \centering
    \includegraphics[width=0.45\textwidth]{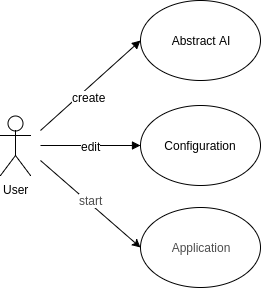}
    \caption[Pseudo-\acrshort{uml} representation of the researcher interacting with the software]{Pseudo-\acrshort{uml} representation of the researcher interacting with the software}
    \label{fig:researcher}
\end{figure}

One of the most important things the researcher has to do manually will be to train an abstract \acrshort{ai} that is needed to sort out the downloaded pictures. In order to do that, a small dataset needs to be created by hand and given into a framework. Afterwards, the configuration must be edited to point to the right path(s) and files. Also, the developer needs to specify the categories the actual \acrshort{ai} will be trained on in the configuration, so that a crawler can automatically download the images later on.

Besides those already described (and shown in figure \ref{fig:researcher}), no further steps should be required. Therefore, the interaction between researcher and software are still very limited, even though there are some manual steps to be done. The consequential system and data requirements are discussed in the next subsections.

\subsection{System requirements}

In the user requirements, there were already traced out a few demands for the system. Of course, the application is more complex and it is necessary to define more conditions for a fully working software.  The upcoming subsections shall give a detailed analysis of the functional requirements for different parts of the system.

\subsubsection{Neural network}

As outlined earlier in \ref{sssec:deeplearning}, the chosen \acrlong{nn} defines the error rate accomplished by the \acrlong{ai}. Because of this, a network with an excellent top-5 and top-1 error rate should be chosen. Also, this \acrshort{nn} needs to be already available as a model for the framework, as this reduces the code needed for the prototypical implementation drastically. 

\subsubsection{Machine learning framework} \label{sssec:framework}

The framework used to assemble the \acrlong{ai} is one of the most critical external software-tools for this approach. The one to be used must match the following criteria:

\begin{itemize}
    \item It (obviously) is capable of creating an \acrshort{ai} for image processing
    \item It is well documented, so that it is easy to learn how to use it
    \item It is implemented so that it makes use of one or multiple \acrshort{gpu}(s), otherwise it would be too slow when working with a large amount of images
    \item It is developed continuously, using state of the art mechanisms and methods
    \item It provides an image for virtualization, so that the process of setting up the system can be automated and further development may include different environments with the same base
    \item It supplies a easily usable binding/wrapping for a programming language the developer is familiar with, in order to flatten the learning curve and saving time
    \item It performs well on the chosen \acrlong{nn}
\end{itemize}

\subsubsection{Crawler} \label{sssec:crawler}

To be able to decide whether to program an own crawler or to use an existing one, it should be evaluated if there already is one capable of exactly what is needed. If so, no time should be invested in reinventing the wheel. An acceptable crawler must include the following functionalities:

\begin{itemize}
    \item It is a library written in the same programming language the framework uses
    \item It can search and download pictures from (multiple) search engines
    \item It is implemented to use multiple threads to give a better performance and download multiple images at once
    \item Its download mechanism is extendable, so that operations can be carried out on the downloaded files while the queue is completed 
\end{itemize}

\subsubsection{Database}\label{sssec:db}

As the state of the contextual object data describing a taxon (not the dataset/images) should be saved persistently, too, a \acrfull{db} is required. Because the relations between single \acrshort{db} entries are perhaps quite minimal and the \acrlong{db} may also be used as a caching system for temporary data, a \acrfull{sql} \acrshort{db} might be too sedate in this case. Also, a mapping of (taxonomic) hierarchies in relational \acrlongpl{db} can quickly become inefficient when relying on traditional methods without using nested sets, for example. Instead, a No\acrshort{sql} \acrshort{db} should come to use. This type of \acrshort{db} is able to store objects (converted to strings) directly without having to dissect them explicitly. Some NoSQL solutions can even run in the machine's memory, hence they can provide results a lot faster than traditional \acrlongpl{db}, which tend to read each entry from the \acrfull{hdd}. 

\subsubsection{Hardware} \label{sssec:hardware}

The computer on which the implementation will be developed and evaluated on has to fulfill a few prerequisites, too. In machine learning, the most crucial part in order to get a result in an endurable time is the \acrshort{gpu}. Without it, all work will be shifted onto the \acrfull{cpu}, which is significantly slower in most cases involving complex mathematical calculations, e.g. training an \acrshort{ai} for image processing could take weeks instead of hours. That being said, there must be a modern, powerful \acrshort{gpu} available for a high load of processing.

For the crawling and downloading of multiple images at a time, a \acrshort{cpu} capable of handling several threads is needed. Furthermore, the internet connection is also a factor of concern. These two components have to be consistent with each other. If the internet connection is too slow, the \acrshort{cpu} will be running idle, the other way around the bandwidth cannot be maxed out. Hence, both of them should perform well according to today's standards.

Another thing which could limit the speed and performance is the hard drive. Not so much during the download, but rather while training the \acrlong{nn}. In that phase, thousands of images will be read from the drive, creating many random access operations. Because \acrfullpl{ssd} are much faster than the "old" \acrshortpl{hdd}, as benchmarks have shown\footnote{\url{http://www.tomshardware.com/reviews/ssd-upgrade-hdd-performance,3023-6.html}}, when doing this kind of operations, the computer should be built with a \acrshort{ssd}.

The last thing to mention in this subsection would be the memory, or \acrfull{ram} as it is called officially. Because a large amount of data (like the cache, as described in \ref{sssec:imgrec}) will be loaded into it and it is not really a bottleneck when looking at the machine's overall performance, the memory's size is more relevant than its speed. Thus, it should have enough capacity to store the \acrshort{db}'s content plus additional space for handling other operations.

\subsection{Data requirements}

Of course, there are also requirements for the data of which the training set(s) shall consist. Because there will be two different datasets created (one manually, one automatically), each must meet distinct conditions. 

To train the abstract \acrshort{ai} used for picture validation, only a small dataset with two categories is needed. But collecting images will still take some time and because of this it would be wise to choose a good source which can deliver many images of a certain quality. In the best case, the images are already validated, so that the researcher only needs to bring them in the right format for the chosen framework.

For the automatically collected images in the other dataset, there are almost no specific requirements, besides one: The file type. In order to automatically download the images and convert them to make them usable for the framework, they must be converted into a certain format, like the one standardized by the \acrfull{jpeg}, e.g. 

During the evaluation, there is a third, independent dataset needed, which must not contain any items from the one used for training, because this would alter the results. Instead, a small subset of the training data could be taken away from it and be used to create a validation set.
\\
\\
\\
The taxonomic data does not have to be provided in a special format, as the import of this data needs to be solved programmatically, anyway. It would be easier to fetch it through an \acrshort{api} instead of downloading it manually and parsing it afterwards, but this should be only an optional requirement.

%% file: kapitel/5_conception.tex
\section{Conception}\label{conception}

During the conception, the current process of creating a dataset for the training of a \acrlong{nn} will be described and a suggestion of how it could be changed in favor of automation will be given. Afterwards, a software for each purpose described before needs to be chosen by evaluating different candidates. At the end, a proposal for the architecture of the prototypical implementation shall be presented.

\subsection{Description of the process} \label{ssec:pdesc}

\begin{figure}[htbp]
    \centering
    \includegraphics[width=1.0\textwidth]{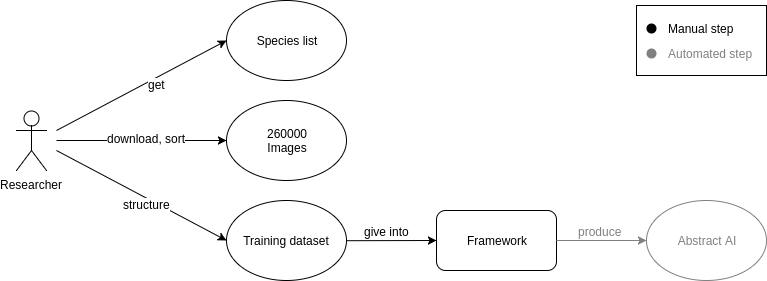}
    \caption[Manual process]{Manual process}
    \label{fig:process_manual}
\end{figure}

\begin{figure}[htbp]
    \centering
    \includegraphics[width=1.0\textwidth]{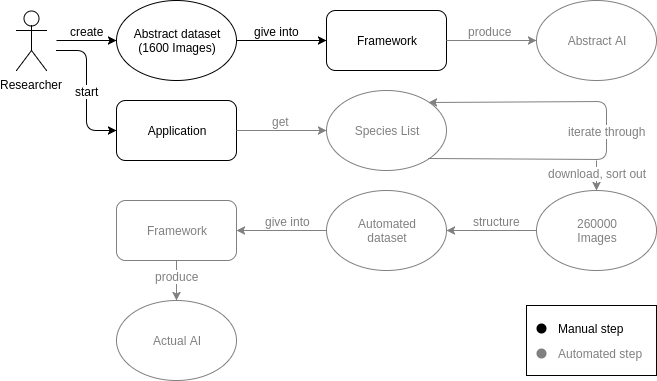}
    \caption[Automated process]{Automated process}
    \label{fig:process_auto}
\end{figure}

The following example shall clarify how the proposed approach can help a scientist by going through the process of creating a dataset step-by-step. Figure \ref{fig:process_manual} and \ref{fig:process_auto} show a schematic representation of the respective process.
\\
\\
Use case: A scientist wants to train an \acrshort{ai} for differentiating images of species under the taxonomic order "Anseriformes" (which contains species such as ducks, geese and relatives).
\\
\\
\textbf{Pre-automation}

\begin{enumerate}
    \item Get a list of all species the order "Anseriformes" includes
    \item Gather images for each of the 325 species. For a good detection there should be between 600 and 1000 pictures per category (as suggested by an exemplary dataset\footnote{\url{http://download.tensorflow.org/example_images/flower_photos.tgz}}), so in this case there would be needed \textasciitilde800*325 = 260000 images.
    \item Sort images out manually 
    \item Convert them into the right file type for the framework
    \item Bring the dataset into the framework's required format (folder structure, etc.)
    \item Let the framework train the \acrshort{ai} based on the acquired dataset
\end{enumerate}

\newpage
\textbf{Post-automation}

\begin{enumerate}
    \item Train an abstract \acrshort{ai} which only knows 2 categories: "bird" and "not bird":
    \begin{enumerate}
        \item Gather images for 2 categories (only 1600 pictures instead of 260000)
        \item Let the framework train the abstract \acrshort{ai}
    \end{enumerate}
    \item \textit{Programmatically} get a list of all species the order "Anseriformes" includes (this could be done by making use of an existing \acrshort{api})
    \item \textit{Programmatically} iterate through the list and crawl/download images from a search engine for each species
    \item \textit{Programmatically} sort out images using the abstract \acrshort{ai} mentioned above
    \item \textit{Programmatically} let the framework train the actual \acrshort{ai} based on the acquired dataset
\end{enumerate}

\subsection{Image classification networks}

For image classification, there have been developed several specialized \acrshortpl{nn}, such as \textit{AlexNet} \cite{alexnet}, \textit{VGG} \cite{vgg}, \textit{OverFeat} \cite{overfeat} or \textit{GoogleNet} \cite{googlenet} (also called \textit{Inception} \cite{inceptionv3} later on), of which some even surpassed human recognition abilities on datasets like CIFAR\footnote{\url{https://www.cs.toronto.edu/~kriz/cifar.html}}, MNIST\footnote{\url{http://yann.lecun.com/exdb/mnist/}} or the ImageNet, which is used in the \acrshort{ilsvrc}.

The decision of which network to choose becomes quite simple when focusing only on their error rates. The third version of Inception gave the best results to date in both top-5 and top-1 error rates, achieving "\textit{3.5\% top-5 error on the validation set [...] and 17.3\% top-1 error on the validation set}", as reported by its creators.

\subsection{Choosing the framework}

Based on the criteria from \ref{sssec:framework}, an \acrshort{ai}-framework has to be chosen. But first, the most commonly used ones shall be described to create an impression of where they originated from and how they work. The most difficult part of this is staying up-to-date, because most of the frameworks have swift development cycles and a comparison from half a year ago could be already outdated, as flaws may have been turned into advantages.

Instead of writing down a subjective impression for every individual framework, they will be presented by a quotation from their official description. This way, there is no opinion imprinted from the author onto the reader.

\subsubsection{Available options}

\textbf{Caffe (v1.0)}\footnote{\url{http://caffe.berkeleyvision.org/}}

\begin{quotation}
"\textit{Caffe provides multimedia scientists and practitioners with a clean and modifiable framework for state-of-the-art deep learning algorithms and a collection of reference models. The framework is a BSD-licensed C++ library with Python and MATLAB bindings for training and deploying general-purpose convolutional neural networks and other deep models efficiently on commodity architectures. [...] Caffe is maintained and developed by the Berkeley Vision and Learning Center (BVLC) with the help of an active community of contributors on GitHub.}" -- quotation from Caffe's whitepaper \cite{caffe}.
\end{quotation}

\begin{figure}[htbp]
    \centering
    \includegraphics[width=1.0\textwidth]{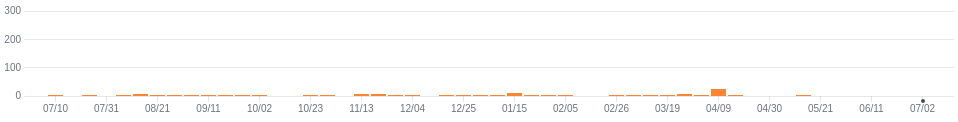}
    \caption[Contributions in the Caffe repository on Github]{Contributions in the Caffe repository on Github}
    \label{fig:commits_caffe}
\end{figure}

\textbf{Torch (v7)}\footnote{\url{http://torch.ch/}}

\begin{quotation}
"\textit{Torch7 is a versatile numeric computing framework and machine learning library that extends Lua. Its goal is to provide a flexible environment to design and train learning machines. Flexibility is obtained via Lua, an extremely lightweight scripting language. High performance is obtained via efficient OpenMP/SSE and CUDA implementations of low-level numeric routines. [...] With Torch7, we aim at providing a framework with three main advantages: (1) it should ease the
development of numerical algorithms, (2) it should be easily extended (including the use of other
libraries), and (3) it should be fast.}" -- quotation from Torch7's whitepaper \cite{torch7}.
\end{quotation}

\begin{figure}[htbp]
    \centering
    \includegraphics[width=1.0\textwidth]{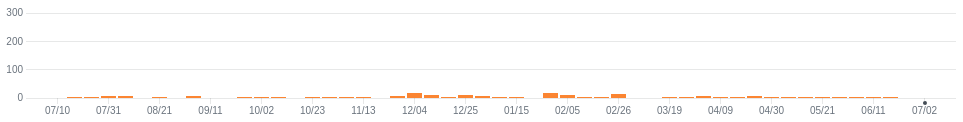}
    \caption[Contributions in the Torch repository on Github]{Contributions in the Torch repository on Github}
    \label{fig:commits_torch}
\end{figure}

\newpage
\textbf{TensorFlow (v1.3)}\footnote{\url{https://www.tensorflow.org/}}

\begin{quotation}
"\textit{TensorFlow is an interface for expressing machine learning algorithms, and an implementation for executing such algorithms. A computation expressed using TensorFlow can be executed with little or no change on a wide variety of heterogeneous systems, ranging from mobile devices such as phones and tablets up to large-scale distributed systems of hundreds of machines and thousands of computational devices such as GPU cards. The system is flexible and can be used to express a wide variety of algorithms, including training and inference algorithms for deep neural network models [...], including speech recognition, computer vision, robotics, information retrieval, natural language processing, geographic information extraction, and computational drug discovery}" -- quotation from TensorFlow's whitepaper \cite{tensorflow}.
\end{quotation}

\begin{figure}[htbp]
    \centering
    \includegraphics[width=1.0\textwidth]{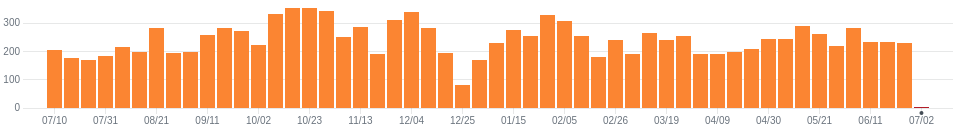}
    \caption[Contributions in the TensorFlow repository on Github]{Contributions in the TensorFlow repository on Github}
    \label{fig:commits_tensorflow}
\end{figure}

\textbf{Neon (v2.0.0)}\footnote{\url{https://www.intelnervana.com/neon/}}

\begin{quotation}
"\textit{neon is Intel Nervana ‘s reference deep learning framework committed to best performance on all hardware. Designed for ease-of-use and extensibility.}
\\
\\
\textit{Features include:}

\begin{itemize}
    \item \textit{Support for commonly used models including convnets, RNNs, LSTMs, and autoencoders. [...]}
    \item \textit{Tight integration with our state-of-the-art GPU kernel library}
    \item \textit{3s/macrobatch (3072 images) on AlexNet on Titan X (Full run on 1 GPU ~ 32 hrs)}
    \item \textit{Basic automatic differentiation support}
    \item \textit{Framework for visualization}
    \item \textit{Swappable hardware backends: write code once and deploy on CPUs, GPUs, or Nervana hardware}" -- quotation from Neon's documentation\footnote{\url{http://neon.nervanasys.com/docs/2.0.0/}}.
\end{itemize}
\end{quotation}

\begin{figure}[htbp]
    \centering
    \includegraphics[width=1.0\textwidth]{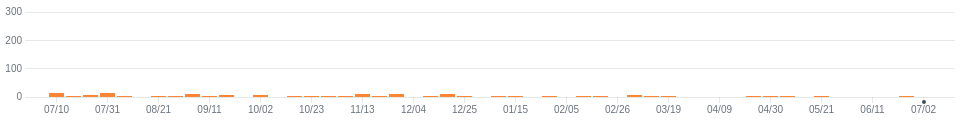}
    \caption[Contributions in the Neon repository on Github]{Contributions in the Neon repository on Github}
    \label{fig:commits_neon}
\end{figure}

\textbf{Theano (v0.9.0)}\footnote{\url{http://www.deeplearning.net/software/theano/}}

\begin{quotation}
"\textit{Theano is a Python library that allows to define, optimize, and evaluate mathematical expressions involving multi-dimensional arrays efficiently. Since its introduction it has been one of the most used CPU and GPU mathematical compilers – especially in the machine learning community – and has shown steady performance improvements. Theano is being actively and continuously developed since 2008, multiple frameworks have been built on top of it and it has been used to produce many state-of-the-art machine learning models.}" -- quotation from Theano's whitepaper \cite{theano}.
\end{quotation}

\begin{figure}[htbp]
    \centering
    \includegraphics[width=1.0\textwidth]{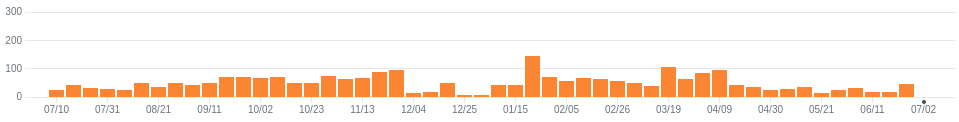}
    \caption[Contributions in the Theano repository on Github]{Contributions in the Theano repository on Github}
    \label{fig:commits_theano}
\end{figure}

\textbf{Deeplearning4j (v.0.9.1)}\footnote{\url{https://deeplearning4j.org/}}

\begin{quotation}
"\textit{Deeplearning4j is an open-source, distributed deep-learning project in Java and Scala spearheaded by the people at Skymind, a San Francisco-based business intelligence and enterprise software firm. [...] Deeplearning4j is distinguished from other frameworks in its API languages, intent and integrations. DL4J is a JVM-based, industry-focused, commercially supported, distributed deep-learning framework that solves problems involving massive amounts of data in a reasonable amount of time. It integrates with Kafka, Hadoop and Spark using an arbitrary number of GPUs or CPUs, and it has a number you can call if anything breaks.}" -- quotation from Deeplearning4j's website\footnote{\url{https://deeplearning4j.org/about}}\textsuperscript{,}\footnote{\url{https://deeplearning4j.org/compare-dl4j-torch7-pylearn}}.
\end{quotation}

\begin{figure}[htbp]
    \centering
    \includegraphics[width=1.0\textwidth]{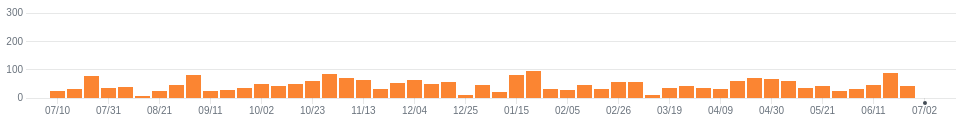}
    \caption[Contributions in the Deeplearning4j repository on Github]{Contributions in the Deeplearning4j repository on Github}
    \label{fig:commits_deeplearning4j}
\end{figure}

\textbf{Summary and explanation} \\
\\
After reading the preceding quotes and their sources, it becomes clear that each framework has its own advantages and drawbacks, depending on the prospect it was created for. For example, while neon is clearly focused on speed, Deeplearning4j's aims to be distributable, but both are made for companies which have to process massive amounts of data on device clusters. 

The graphic below each quote represents a statistic of contributions to the respective framework's source code between 07/02/2016 and 07/02/2017. The chart can be found on every project's github page under the \textit{insights} section, following this scheme: \urlstyle{same}\url{https://github.com/[organisation]/[project]/graphs/commit-activity}\urlstyle{tt}. For torch7, this would result in \url{https://github.com/torch/torch7/graphs/commit-activity}, e.g. Because each graph is scaled to fit the project's maximum value by default, the ones shown in this thesis have been modified to fit the overall maximum value (353). 

Table \ref{table:frameworks} (p. \pageref{table:frameworks}) shall give a resume of how each of the frameworks matches the requirements listed in \ref{sssec:framework}. It was collated using multiple sources\footnote{\url{https://en.wikipedia.org/wiki/Comparison_of_deep_learning_software}}\textsuperscript{,}\footnote{\url{https://blog.paperspace.com/which-ml-framework-should-i-use/}}\textsuperscript{,}\footnote{\url{https://github.com/zer0n/deepframeworks/blob/master/README.md}}\textsuperscript{,}\footnote{\url{https://github.com/soumith/convnet-benchmarks}}\textsuperscript{,}\footnote{\url{https://hub.docker.com/}, searching for each framework} \cite{speedacc}, \cite{compstudy} and the aforementioned statistical graphs. 

\newcolumntype{g}{>{\columncolor[gray]{.9}} m{2cm} }
\newcolumntype{h}{>{\centering\arraybackslash}  m{1.75cm} }
\begin{table}[htbp]
\begin{center}
\begin{tabular}{ |g|h|h|h|h|h|h| }
    \hline
    \rowcolor[gray]{.9} Require-ment & Caffe & Torch7 & Tensor-Flow & Neon & Theano & Deep-learning4j \\ \hline
    GPU support (multiple) & Yes (Yes) & Yes (Yes) & Yes (Yes) & Yes (Cloud only) & Yes (Yes) & Yes (Cloud only) \\ \hline
    Documen-tation, examples \& models & Good & Good & Excellent & OK & OK & Good \\ \hline
    Develop-ment & OK & OK & Excellent & OK & Good & Good \\ \hline
    Official virtualization images & Yes & No & Yes & No & No & Yes, but outdated \\ \hline
    Language bindings & Python, MATLAB & Lua, C & Python, C/C++, Java, Go, R & Python & Python & Java, Scala, Clojure, Python \\ \hline
    Perfor-mance & OK & Good & Good & Excellent & Good & Low \\ \hline
\end{tabular}
\caption[Framework requirement match overview]{Framework requirement match overview (Scale: Excellent > Good > OK > Low > Bad)}
\label{table:frameworks}
\end{center}
\end{table}

The table does not list whether a framework provides a model for \textit{Inception v3} or not, because most of the frameworks do so and if not, there is a tool available to convert them from another frameworks format, like \textit{caffe2neon}\footnote{\url{https://github.com/NervanaSystems/caffe2neon}}, e.g. The same applies for the ability to create an \acrshort{ai} for image processing, as all frameworks are only made for this task.

\subsubsection{Result} \label{sssec:fwres}

As table \ref{table:frameworks} suggests, Tensorflow is the best choice for the approach. It fulfills all requirements and as a bonus, it provides an example Inception-model/\acrshort{nn} for image classification where only the last layer needs to be re-trained\footnote{\url{https://www.tensorflow.org/tutorials/image_retraining}}. This saves a lot of time and will comprehend the performance handicap in comparison to neon.

TensorFlow seems to be not only a good choice for the implementation in this thesis, but is also very popular among developers and researchers, as one can see when looking at the number of stars it has gotten on Github. Each of these stars represents one person following the project's updates.

\begin{figure}[htbp]
    \centering
    \includegraphics[width=1.0\textwidth]{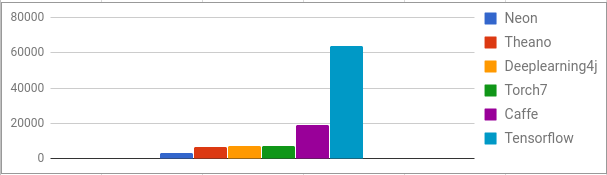}
    \caption[Framework stars on Github]{Framework stars on Github (07/16/2017)}
    \label{fig:framework_graph}
\end{figure}

Because of the many people contributing to and using the project, more examples and use cases can be provided, which in turn attracts new users, and so on. Arguing that Tensorflow is backed by a corporation (Google) does not say much about the framework's quality, because Torch7 (Facebook, Twitter, Google, Yandex, etc.)  and neon (Intel) are also funded by companies, but do not seem to be as successful.

\subsection{Crawler}

There are already several crawlers capable of multi-threading and downloading images from search engines, as described in \ref{sssec:crawler}. When searching for "\textit{python search engine image crawler multi thread}" (python will be the employed programming language) on Google, the first result is \textit{icrawler}\footnote{\url{https://github.com/hellock/icrawler}}. icrawler seems to do everything required: It is capable of multi-threading, can query different search engines, is easily extendable and written in python. Also, it seems to be steadily developed and provides examples for many different use cases.

All other image crawlers listed in the search results are abandoned and have not been updated in the last two years at the time of searching. Also, most of them are not designed to be integrated into other applications\footnote{\url{https://github.com/sanghoon/image_crawler}}\textsuperscript{,}\footnote{\url{https://github.com/sananth12/ImageScraper/}}\textsuperscript{,}\footnote{\url{https://github.com/NikolaiT/GoogleScraper}}.

Because of these circumstances, icrawler was chosen to be the designated crawler to be used in the prototypical implementation.

\subsection{Database}

NoSQL data management tools have become more and more popular over the last years. A very well-known project in this field is \textit{redis}\footnote{\url{https://redis.io/}}. On its homepage, it is described as a "\textit{in-memory data structure store, used as a database, cache and message broker}", which means it stores data as key-value pairs inside a machine's memory to be as fast as possible. Because of this architecture, redis it is way ahead of other NoSQL solutions, such as \textit{MongoDB}\footnote{\url{https://www.mongodb.com/}}, for example, in terms of performance\footnote{\url{http://badrit.com/blog/2013/11/18/redis-vs-mongodb-performance}}.

To achieve data persistence, it writes a backup on the \acrshort{hdd} in a pre-defined intervall and when it is shut down. It has bindings for almost every modern programming language and in combination with \acrshort{json} or other serialization techniques, it is possible store all types of objects in it easily. Listing \ref{lst:redis} shows an example of getting and settings values from and to redis in python.

\begin{minipage}{\linewidth}
\lstset{language=python}
\begin{lstlisting}[frame=htrbl, caption={Getting and setting values in redis}, label={lst:redis}]
import redis
r = redis.Redis()
r.set('test', 'Hello World!') # returns: True
r.get('test') # returns: 'Hello World!'
\end{lstlisting}
\end{minipage}

In conclusion, redis is exactly what was described as a requirement in \ref{sssec:db} and therefore will be used as a cache and \acrlong{db}.

\subsection{API} \label{ssec:api}

Using an \acrlong{api} is an optionial requirement, but as there already are some for taxonomic entries, one of them should be used. The \acrfull{gbif}\footnote{\url{http://www.gbif.org/}} provides a taxonomy \acrlong{api}\footnote{\url{http://www.gbif.org/developer/species}} which can be used freely without any authentication or request-quotas, so that it seems to be ideal for this task.

Sending a GET \acrshort{http}-request to \url{http://api.gbif.org/v1/species/1108/children?limit=500} will result in 21 families, notated in \acrshort{json}, as listing \ref{lst:gbif} shows exemplary. \textit{1108} in the \acrshort{url} is the identifier for "\textit{Anseriformes}", following the example described before. Querying the same endpoint with the keys from the result will return their children, and so on. That means if used recursively, the \acrshort{api} can provide information about all taxa present under a specified identifier.

\begin{minipage}{\linewidth}
\lstset{language=python}
\begin{lstlisting}[frame=htrbl, caption={Result of a HTTP GET request to the GBIF's API}, label={lst:gbif}]
{
    offset: 0,
    limit: 100,
    endOfRecords: true,
    results: [
    {
        key: 2986, 
        [...]
        scientificName: "Anatidae",
        [...]
        rank: "FAMILY",
        [...]
    }, [...] ]
\end{lstlisting}
\end{minipage}

As an \acrshort{api} will be made use of, there is a need for an interface able to send \acrshort{http}-requests. There are several libraries written in python which can do this. Listing \ref{lst:urllib} (\textit{urllib}), \ref{lst:httplib} (\textit{httplib}) and \ref{lst:requests} (\textit{requests}\footnote{\url{python-requests.org}}) show example code for doing a GET request in python3.

\begin{minipage}{\linewidth}
\lstset{language=python}
\begin{lstlisting}[frame=htrbl, caption={Example GET request with urllib}, label={lst:urllib}]
import urllib.request
connection = urllib.request.urlopen('http://api.gbif.org/v1/species/1')
response = connection.read()
print(response)
\end{lstlisting}
\end{minipage}

\begin{minipage}{\linewidth}
\lstset{language=python}
\begin{lstlisting}[frame=htrbl, caption={Example GET request with httplib}, label={lst:httplib}]
import http.client
connection = http.client.HTTPConnection("api.gbif.org")
connection.request("GET","/v1/species/1")
response = connection.getresponse().read()
print(response)
\end{lstlisting}
\end{minipage}

\begin{minipage}{\linewidth}
\lstset{language=python}
\begin{lstlisting}[frame=htrbl, caption={Example GET request with requests}, label={lst:requests}]
import requests
response = requests.get('http://api.gbif.org/v1/species/1')
print(response.text)
\end{lstlisting}
\end{minipage}

Requests is the most high-level interface and therefore the easiest to use. Also, it implements several things like Cookies or \acrshort{json}, which the others can not handle themselves. Thus, requests will come to use.

\subsection{Virtualization} \label{ssec:docker}

The topic of virtualization was only given a small amount of attention until now, because the proposed system described before could of course also work without being run inside a isolated environment. But still, virtualization provides a lot of automation potential, security advantages, scalability and protability. Because of these factors, which might become even more important, if the system is adopted by the biodiversity warehouse, the prototype will already make use of it.

Currently, the state of the art technology to provide a virtual environments is Docker\footnote{\url{https://www.docker.com/}}. With Docker, all installation and configuration steps can be defined in a so called \textit{Dockerfile} and thus become automated. Using this file, an \textit{image} representing the desired software stack is built. Images can inherit all steps (also called \textit{layers}) from other (parent) images. They can be distributed to other developers and are started inside a \textit{container}. Docker containers can communicate with each other (and their host machine, of course), but are treated as if each of them would be run on another device. Even though Docker is a solution for virtualization and includes all of its benefits, the usual need for more resources is omitted, because it can run processes natively on Linux without creating an overhead. This is the reason Docker became so popular among software developers. For more detailed information, please see Docker's documentation\footnote{\url{https://www.docker.com/what-docker}}\textsuperscript{,}\footnote{\url{https://docs.docker.com/}}.

If more than one container needs to be run in order to provide a working system, this can be done by using Docker-Compose\footnote{\url{https://docs.docker.com/compose/}}. While Docker alone is also capable of doing so, Docker-Compose simplifies this task by reading a configuration from the \textit{docker-compose.yml} file. This file defines the images to be started, environment variables, which containers can communicate with each other and many other things\footnote{\url{https://docs.docker.com/compose/compose-file/}}. To clarify this, listing \ref{lst:docker} shows the commands needed by Docker to build an image with a Dockerfile, run it afterwards and let it communicate to a container running a redis-image. Listing \ref{lst:docker-compose} shows the content of a docker-compose.yml file defining the same task, but only needs one (short) command to be started: \textit{docker-compose up --build}

\begin{minipage}{\linewidth}
\lstset{language=bash}
\begin{lstlisting}[frame=htrbl, caption={Docker commands}, label={lst:docker}]
docker pull redis
docker build --tag custom_tensorflow:latest ./docker/tensorflow/
docker run --name redis redis
docker run --interactive --tty --link redis custom_tensorflow
\end{lstlisting}
\end{minipage}

\begin{minipage}{\linewidth}
\begin{lstlisting}[frame=htrbl, caption={Example docker-compose.yml}, label={lst:docker-compose}]
version: '3'
services:
  redis:
    image: redis
  tensorflow:
    build:
      context: ./docker/tensorflow
    stdin_open: true
    tty: true
    links:
      - redis
\end{lstlisting}
\end{minipage}

Even though this example is quite minimalistic (there are no open ports yet, e.g.), one can already see that Docker itself can get confusing very fast and the complexity of the commands needed grows with the options and number of containers. Meanwhile, the command for Docker-Compose stays the same and all modifications are saved persistently in the configuration file. Hence, Docker will be used in conjunction with Docker-Compose for the implementation.

\subsection{Proposed system architecture}

The description of the process after the automation in \ref{ssec:pdesc} serves as a good starting point to create a concept for the prototype's system architecture. Assuming that the abstract \acrshort{ai} has already been trained manually to detect birds on images and can be integrated in the implementation, the workflow of the system should look as follows:

When the program is started, it will first request all taxa of a given rank for a taxonomic entry through an \acrshort{api} reader which queries the species endpoint of the \acrshort{gbif}'s \acrshort{api} recursively, as desrcibed in \ref{ssec:api}. In the example mentioned above, this would give 325 species back as a result for the requested order "\textit{Anseriformes}". Each of these results will then be stored inside the \acrshort{db} (redis) to save it persistently.

Just after the \acrshort{api} reader has finished its task, the crawler begins to query one or more search engines with the entries from the \acrshort{db}. As a result, it will get back links to images of different species in this example. The list of these \acrshortpl{url} will be used to instantiate a built-in downloader insider the crawler, which can only download the pictures to a specified directory by default.

To fit TensorFlow's specifications, the directories need to be named after the categories (species). It is important to mention that the crawler and downloader can't do anything complex by themselves, because at this point, the requirement of an extendable crawler becomes inevitable. It must be extended in order to be capable of communicating with the abstract \acrshort{ai}, which then in turn can tell the downloader whether to keep a downloaded image and save it to the \acrshort{hdd} or delete it if there is no bird in it.

\begin{figure}[htbp]
    \centering
    \includegraphics[width=1.0\textwidth]{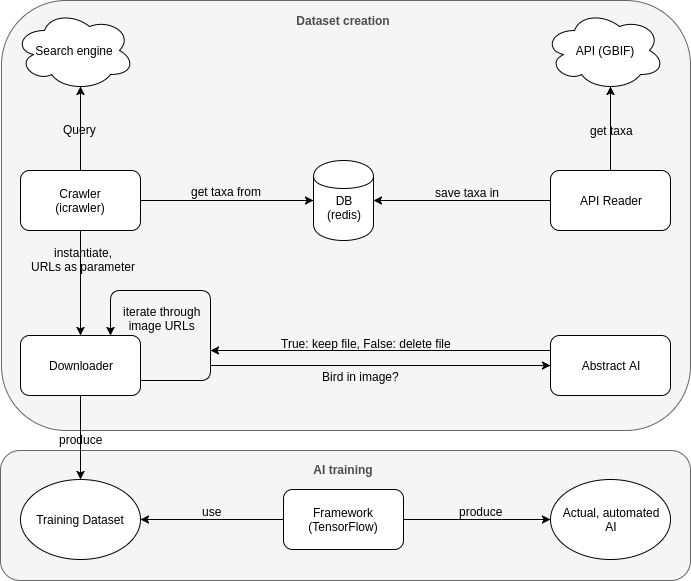}
    \caption[Proposed system architecture]{Proposed system architecture}
    \label{fig:concept_architecture}
\end{figure}

This way, the system will produce a dataset that can be used by TensorFlow to create the actual \acrlong{ai}, which can tell the 325 different species apart. The work flow described before is shown in figure \ref{fig:concept_architecture} (p. \pageref{fig:concept_architecture}, where all components are pictured for a better overview.

As the re-training example mentioned in \ref{sssec:fwres} provides a single command (calling a pre-compiled, binary, executable file with a few arguments) to start the whole training process, it should be seriously considered to make use of it and check the exit status afterwards, instead of re-writing the example in python to eventually achieve the same result. Also, the training needs to be manually started in order to compare the manually collected dataset in the evaluation, so that having a individual command for that case appears to be quite useful.

Usually, the example would need to be compiled first on every machine it shall be run on. Using Docker, this problem can be overcome by simply creating a pre-built image in which the compilation step was already done once. If the development machine is not fast enough or needs to be replaced for any reason, this can save a lot of time. In a productive environment, the image could be started multiple times on different devices, so that the workload can be distributed evenly among them.
\\

%% file: kapitel/6_implementation.tex
\section{Implementation}\label{implementation}

The process of the prototypical implementation shall be described in this section. In doing so, an overview of the development environment will be given and code fragments are to be examined. In addition, problems arising while programming need to be depicted. The given examples are kept as minimalistic as possible, so that they can be understood easier without having to look at their respective context inside the entire source code.

\subsection{Hardware} \label{ssec:impl-hardware}

In \ref{sssec:hardware}, there was a hardware setup outlined which constitutes the required machine specifications. The private computer owned by this thesis' author meets and exceeds these requirements, so that there is no need to buy or rent anything in order to develop the prototype. The following list shall give an overview of the crucial hardware parts:

\begin{description}
\item[\acrshort{gpu}]\hfill \\
\textit{EVGA GeForce GTX 1080 FTW Gaming ACX 3.0}: Mainly produced for gaming, the GTX 1080 can also be employed for tasks in the field of \acrlong{ml}, as it shows good performance when used with image processing \acrshortpl{nn}\footnote{\url{https://github.com/jcjohnson/cnn-benchmarks/blob/master/README.md}}.
\item[\acrshort{cpu}]\hfill \\
\textit{Intel Core i7 6700K}: This processor has 4 cores and runs on a clock rate of 4.0 GHz. It is capable of using 8 threads at once and thus will be able to provide a excellent performance for the crawler.
\item[\acrshort{hdd}]\hfill \\
\textit{Samsung 840 EVO 500GB SSD}: This \acrshortpl{ssd} model can read and write up to more than 500MB per second, which should be enough to accomplish an acceptable training time.
\item[\acrshort{ram}]\hfill \\
\textit{16GB Corsair Vengeance LPX black DDR4-3000 DIMM CL15 Dual Kit}: The name of this kit includes the module specifications: Each of the two provides 8GB of memory. This should make the \acrshort{ram} capable of holding all the data needed for the \acrshort{db} and cache.
\end{description}

With this technical equipment, the development and testing can be done properly and within a decent time duration.

\subsection{Software and setup}

Because of the virtualization through Docker, the \acrfull{os} of the development system is pretty insignificant. But still, it should be mentioned at this point. Its name is \textit{Antergos}\footnote{\url{https://antergos.com/}} and it is build on top of the \textit{Arch Linux}\footnote{\url{https://www.archlinux.org/}} base. The developers of Arch Linux maintain many up-to-date packages in their repositories and thus make installing a software stack very simple. For example, the official Docker documentation for Debian\footnote{\url{https://www.debian.org/}} and Ubuntu\footnote{\url{https://www.ubuntu.com/}} suggests removing the official (outdated) packages and adding a new repository provided by the docker team, which takes several steps. Under Arch, only one command is needed: \textit{sudo pacman -S docker}. The same applies for Docker-Compose and many other tools and therefore saves a lot of time when setting up a development environment.

In Docker, a container is completely isolated from its host, unless the developer speciefies something else. Ports need to be exposed and folders have to be mounted inside Docker volumes. With a volume, files can be written and shared from the container to the host and vice versa. This creates the possibility of storing data persistently, because normally, all data inside a Docker container is lost when it is shut down (not to be confused with being stopped). Docker already provides a \textit{-v} or \textit{--volume} command line argument to create volumes, but Docker-Compose simplifies this principle and lets one define the volumes directly inside the docker-compose.yml file. Listing \ref{lst:volumes} shows an example for redis, enabling it to save its key-value store persistently.

\begin{minipage}{\linewidth}
\begin{lstlisting}[frame=htrbl, caption={Volumes in Docker-Compose}, label={lst:volumes}]
version: '3'
services:
  redis:
    image: redis
    command: redis-server --appendonly yes
    volumes:
      - ./data/db:/data
\end{lstlisting}
\end{minipage}

Docker itself does not support the use of \acrshortpl{gpu} inside of its containers natively. This problem is solved by a manufacturer (Nvidia\footnote{\url{http://www.nvidia.com/page/home.html}}). The company provides a modified version of the software called \textit{nvidia-docker}\footnote{\url{https://github.com/NVIDIA/nvidia-docker}}. Using nvidia-docker, it is possible to mount the host system's graphics driver and -card into a container in order to run calculations through/on it. There is also a tool named \textit{nvidia-docker-compose}\footnote{\url{https://github.com/eywalker/nvidia-docker-compose}}, which modifies a given docker-compose.yml so that Docker-Compose makes use of the aforementioned mounts. This will only work if nvidia-docker was run once before and has created the Docker volume(s) to be mounted. Listing \ref{lst:nvidia-docker-compose} shows a modified docker-compose.yml from the example in \ref{ssec:docker}, which can be started by either nvidia-docker-compose or Docker-Compose.

While the command \textit{docker-compose up} will start \textbf{all} services defined in the docker-compose.yml and show their output, \textit{docker-compose run [SERVICE NAME]} will execute a specified service like an application. This means the devloper can interact with the process and send input to it. The TensorFlow Docker image, for example, only starts a shell by default and waits for input. If the image is started with \textit{up}, there is another command needed to attach to and interact with it, which (depending on the command being \textit{attach} or \textit{exec}) may spawn another process inside the container. When using \textit{run}, not only the specified service is started, but all services depending on or linked with it. Thereby it is ensured, that everything works as expected. Because of this behaviour, it is better to use \textit{run} for images that do not represent a daemon application meant to be run in the background, like a webserver or \acrshort{db}, e.g. 

\begin{minipage}{\linewidth}
\begin{lstlisting}[frame=htrbl, caption={Modified docker-compose.yml}, label={lst:nvidia-docker-compose}]
version: '3'
services:
  redis:
    image: redis
  tensorflow:
    build:
      context: ./docker/tensorflow
    stdin_open: true
    tty: true
    devices:
      - /dev/nvidia0
      - /dev/nvidiactl
      - /dev/nvidia-uvm
      - /dev/nvidia-uvm-tools
    volumes:
      - nvidia_driver_381.22:/usr/local/nvidia:ro
    links:
      - redis
volumes:
  nvidia_driver_381.22:
    external: true
\end{lstlisting}
\end{minipage}

\subsection{Development}

Once the system is roughly set up, the development can be begun and the ideas from the conception can be implemented. In this subsection, parts and functionalities of the prototype are be described in detail based on the source code from the implementation.

\subsubsection{Dockerfile}

When (nvidia-)Docker(-Compose) is installed on the host machine, a Dockerfile must be written in order specify the actual system and install all required applications. As mentioned before in \ref{ssec:docker}, it is possible to inherit from other images, which will then already contain all steps included in their respective Dockerfile. Listing \ref{lst:dockerfile} shows an example where the official TensorFlow image is extended. This way, the image classification example is already compiled when building the image and this step does not have to be done again every time the container is started. 

\begin{minipage}{\linewidth}
\begin{lstlisting}[frame=htrbl, caption={Example Dockerfile}, label={lst:dockerfile}]
FROM tensorflow/tensorflow:1.2.1-devel-gpu-py3
WORKDIR /tensorflow/tensorflow/examples/
RUN bazel build -c opt --copt=-mavx image_retraining:retrain
\end{lstlisting}
\end{minipage}

The Docker registry provides tags, because the parent image can change and the provided environment could become unusable in the next version. A tag is set in the \textit{FROM} instruction (after the image name) and determines a version or release of an image so that someone using it is be able to rely on the inherited steps. In the example, the tag is \textit{1.2.1-devel-gpu-py3}, which points to the developers version of TensorFlow of release 1.2.1 that has GPU support enabled and uses python3. For the prototype in the context of this thesis, the developer's version must be used, because it comes with the source code containing the example which is referenced in the Dockerfile and hence required during the build process. The available tags for an image can be found on the project's Docker Hub page, under the "Tags"-tab\footnote{\url{https://hub.docker.com/r/tensorflow/tensorflow/tags/}}.

\subsubsection{API connection}

As the \acrshort{gbif}'s \acrshort{api} does not require any authentication on most endpoints, it is fairly simple to integrate it into an application. The python code in listing \ref{lst:gbif-request} is everything needed to retrieve and print all children of the taxon with identifier 212 (birds). As stated before, this example code makes use of recursion to be able to iterate through different taxonomic layers.

\begin{minipage}{\linewidth}
\lstset{language=python}
\begin{lstlisting}[frame=htrbl, caption={Example code for \acrshort{api} calls}, label={lst:gbif-request}]
import requests

def get_children(id):
    url = ('http://api.gbif.org/v1/species/' +
          str(id) +
          '/children?limit=99999999')
    response = requests.get(url).json()
    for result in response['results']:
        print(result['scientificName'])
        get_children(result['key'])
        
get_children(212)
\end{lstlisting}
\end{minipage}

\subsubsection{Extending the crawler}

\begin{figure}[htbp]
    \centering
    \includegraphics[width=1.0\textwidth]{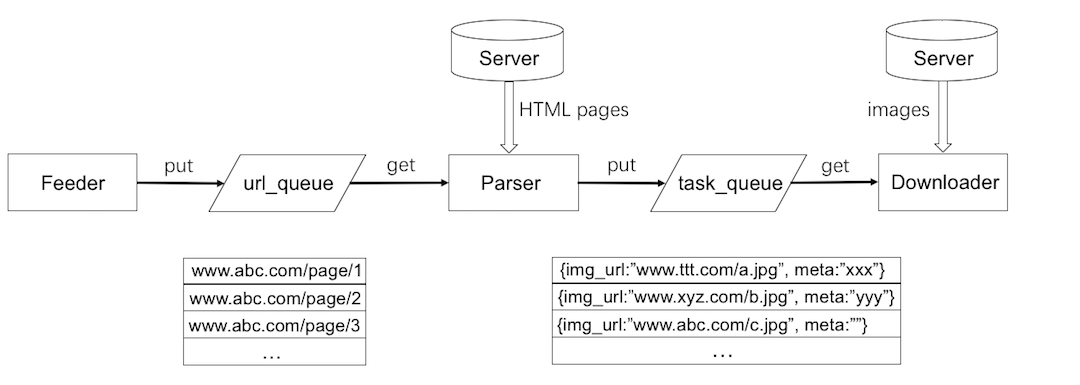}
    \caption[iCrawler architecture]{iCrawler architecture\footnotemark}
    \label{fig:icrawler}
\end{figure}

For a working prototype, the extension of iCrawler is indispensable, because the built-in classes only provide basic functionality. A crawler within the iCrawler context consists of 3 different components (see figure \ref{fig:icrawler}). The feeder "feeds" the parser with \acrshortpl{url} of websites to be parsed. The parser parses the \acrshort{html} of each website and extracts certain elements, such as image \acrshortpl{url}. When it has finished the parsing, the links are passed to the downloader, which will process the queue and download the images.

\footnotetext{\url{http://7xopqn.com1.z0.glb.clouddn.com/workflow.png}}

Besides the default classes from the forestanding figure, iCrawler also includes several built-in implementations for popular search engines. A standard use of iCrawler's \textit{GoogleImageCrawler} is shown in listing \ref{lst:icrawler-google}.

\begin{minipage}{\linewidth}
\lstset{language=python}
\begin{lstlisting}[frame=htrbl, caption={Standard iCrawler code}, label={lst:icrawler-google}]
from icrawler.builtin import GoogleImageCrawler

google_crawler = GoogleImageCrawler(
    parser_threads=2,
    downloader_threads=4
    storage={'root_dir': '/some/path'}
)
google_crawler.crawl(
    keyword='"duck"',
    offset=0,
    max_num=50,
    date_min=None,
    date_max=None,
    min_size=(500, 500),
    max_size=None
)
\end{lstlisting}
\end{minipage}

While the crawler class (GoogleImageCrawler) already does what it should (crawl image links from Google), the corresponding default downloader class (\textit{ImageDownloader}) needs to be extended. Normally, it only downloads and renames images, but for this thesis' approach it has to make use of the abstract \acrshort{ai}. To achieve this, the \textit{download} method has to be re-written in a new class. This class then needs to be given into the existing crawler as an argument, as shown in listing \ref{lst:icrawler-google-ext}.

\begin{minipage}{\linewidth}
\lstset{language=python}
\begin{lstlisting}[frame=htrbl, caption={Extended iCrawler code}, label={lst:icrawler-google-ext}]
from extended.downloader import ExtendedImageDownloader
from icrawler.builtin import GoogleImageCrawler

google_crawler = GoogleImageCrawler(
    parser_threads=2,
    downloader_threads=4
    storage={'root_dir': '/some/path'},
    downloader_cls=ExtendedImageDownloader
)
[...]
\end{lstlisting}
\end{minipage}

This extension support by iCrawler's architecture is very valuable, because the extended class can be re-used for all other built-in crawlers, such as \textit{BaiduImageCrawler} or \textit{BingImageCrawler}, e.g. and does not have to be re-written each time.

\subsubsection{Abstract AI} \label{sssec:abstract_ai}

The abstract \acrshort{ai} is responsible to filter out unwanted pictures. For the example used in this thesis, that would be images containing no birds, but of course, the same approach can be used on every other categorization scheme that can be abstracted.

\begin{minipage}{\linewidth}
\lstset{language=bash}
\begin{lstlisting}[frame=htrbl, caption={Command to train a TensorFlow model}, label={lst:train-abstract-ai}]
cd /tensorflow/bazel-bin/tensorflow/examples/image_retraining/
./retrain  --output_labels /output/bird.txt \
--output_graph /output/bird.pb \
--image_dir /images/bird/
\end{lstlisting}
\end{minipage}

After the abstract \acrshort{ai} was trained on a small manually collected dataset, using the two commands in listing \ref{lst:train-abstract-ai} inside the Docker container, the hereby created model can be used in the prototype. The TensorFlow documentation mentions an example python script\footnote{\url{https://github.com/eldor4do/TensorFlow-Examples/blob/master/retraining-example.py}}, which loads an image and uses an existing model to print out the top-5 guessed labels for it. By modifying this code to work with a dynamic number of results and image path, the integration of an abstract \acrshort{ai} into the prototype can be done  quite easily.

\begin{figure}[htbp]
    \centering
    \includegraphics[width=1.0\textwidth]{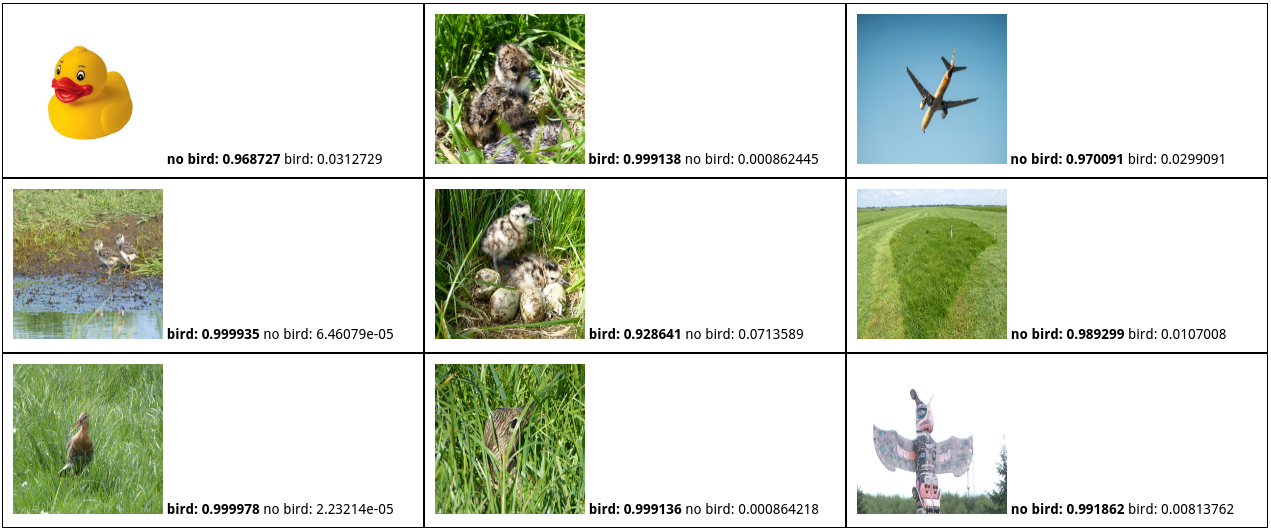}
    \caption[Abstract \acrshort{ai} sample test]{Abstract \acrshort{ai} sample test}
    \label{fig:samples}
\end{figure}

Before this is done, the abstract \acrshort{ai} needs to be tested for functionality on sample images by the researcher. The pictures should contain positive and negative content (birds and no birds) so that both cases can be examined. Figure \ref{fig:samples} shows a partial view of images being categorized by the abstract \acrshort{ai}. 

The images were provided by Arno Schoppenhorst (BUND Landesverband Bremen), Heide-Rose Vatterrott (Hochschule Bremen - City University of Applied Sciences) and gathered from different sources\footnote{\url{http://www.casa-rosa-online.de/WebRoot/Store20/Shops/38191/4E97/2E91/7F2F/52BB/310D/C0A8/29BA/1788/Quietscheente_mini.jpg}}\textsuperscript{,}\footnote{\url{https://static.pexels.com/photos/355957/pexels-photo-355957.jpeg}}.

\subsection{Problems}

When implementing the concept practically, several problems that were not considered beforehand appear. They are addressed on the following pages and a solution for each of them will be discussed.

\subsubsection{Image size and type} \label{sssec:size}

The search engines' crawler(s) may provide and option to set the maximum file size, which could be used to save disk space. But using it would result in a smaller image set and pictures with positive content could be filtered out before they can be examined by the abstract \acrshort{ai}. To avoid this, the images need to be downloaded, examined and (if the content is positive) resized to a maximum of 500x500 pixels inside the prototype. This will not affect the performance (because TensorFlow's example sizes them internally, anyway) and may lead to better results and a lot of saved disk space.

Something else to look at while downloading the images is their file type. This does not mean in any way that other file types than the one the framework needs can't be downloaded, a conversion was already mentioned in \ref{ssec:pdesc}. But some types, like an animated \acrfull{gif} can not be converted, because it consists of multiple frames, of which some could be empty and need to be filtered out. This can be done by using python's built-in functionality to determine the type of an image\footnote{\url{https://docs.python.org/3.6/library/imghdr.html}} and a library called \textit{wand}\footnote{\url{http://docs.wand-py.org/en/0.4.4/guide/sequence.html}} which is already used to resize downloaded images.

\subsubsection{Duplicate image content}

When employing multiple search engines at once, the images may be downloaded twice or contain the same content, even if they differ in size or saturation. This would not be the case if only Google was used, e.g., because a single engine usually already filters out same images in its search results.

The proposed solution for this problem can be divided into two steps. First, the \acrshort{url} of each image should be saved into the \acrshort{db}, so that a trivial duplication check of the links can be done when crawling the next search engine. In the second step, a so called \textit{average hash} \cite{avghash} needs to be computed for the image. This hashing algorithm does not calculate the result based on every pixel and its color in the image, but rather finds a mean value which can be used to compare similar images. This means that if the resulting hash is the same for two images, their content has to be very alike. There already is a python library\footnote{\url{https://github.com/JohannesBuchner/imagehash}} providing exactly this functionality. The computed hash will then also be saved persistently and compared for each new image to eliminate duplicate images in the dataset.

\subsubsection{Extinct species}

Another problem is that the \acrshort{gbif}'s \acrshort{api} does not only list species that can currently be found (and photographed), but also those which went extinct. This would not be a problem if the status could be determined through the \textit{/children} endpoint, but this is not the case. Instead, the \textit{/speciesProfiles} endpoint has to be requested for each species ID. Because different sources can define a different status (alive or extinct), all sources must be iterated and the higher weighted status has to be accepted. This doubles the amount of \acrshort{http} requests to the \acrshort{api}, but cannot be avoided.

\subsubsection{Taxonomic layer}

When beginning to conceptualize the prototype, the idea was to create an \acrshort{ai} that can distinguish between species on images. But first test-runs showed that this generated two problems: There were too many categories (15758 bird species were retrieved from the \acrshort{api}) and too few images for each of them. This resulted in a very high error rate, which lead to the conclusion that a category should be represented by a higher taxonomic rank. Therefore, the prototype was modified to save the images of a species under its corresponding order. By doing this, the amount of categories was reduced to 40 and there were enough pictures for each one.

Because of this modification, the \acrshort{db} entries need to be formatted in another manner. The entry for each species must include the order it belongs to, so that the downloader can put the images into the right directory, as the folder name is used by TensorFlow to determine the category.

\subsubsection{Progress status}

During the development and testing of the prototype, a need for a visualization of the current progress status emerged from the circumstance that there was no \acrlong{ui} and the first tests took a lot of time. To tackle this problem, a logging method needs to be added to the main class of the prototype, which calculates the remaining time and percentage of successfully downloaded species. This information will then be printed out along with the elapsed time, estimated remaining time and absolute species count. Listing \ref{lst:logging} contains an exemplary output.

\begin{minipage}{\linewidth}
\lstset{language=bash}
\begin{lstlisting}[frame=htrbl, caption={Example logging output}, label={lst:logging}]
2017-06-30 22:24:34,043 - INFO - Crawler
11817/15758 (74.99%), Elapsed: 1 day, 1:38:21, ETA: 8:32:52
\end{lstlisting}
\end{minipage}

As the output in the preceding listing shows, the crawling of the dataset takes almost 1,5 days. In \ref{ssec:impl-hardware} it was mentioned that the development-machine is also the private home computer of the author. This means it is also used for other tasks while crawling the images in the background and can not be left running all day and night, because the energy costs would increase dramatically. Because of this, there is a need to persist the current state when turning off the device. This goal can be reached by saving the current species count into the \acrshort{db}, so that the iteration can continue on the next application start. Persistent states also help recovering from an unlikely external event like a power outage, e.g.

The the training progress of a neural network can be visualized in two ways when using TensorFlow. The retraining example has its own output which logs the training steps as well as the created bottlenecks. "\textit{'Bottleneck' is an informal term we often use for the layer just before the final output layer that actually does the classification.}" - quoted from the example's description. These bottlenecks should be kept persitently with the help of a Docker volume, so that they can be re-used in order to decrease the training time.

\begin{figure}[htbp]
    \centering
    \includegraphics[width=1.0\textwidth]{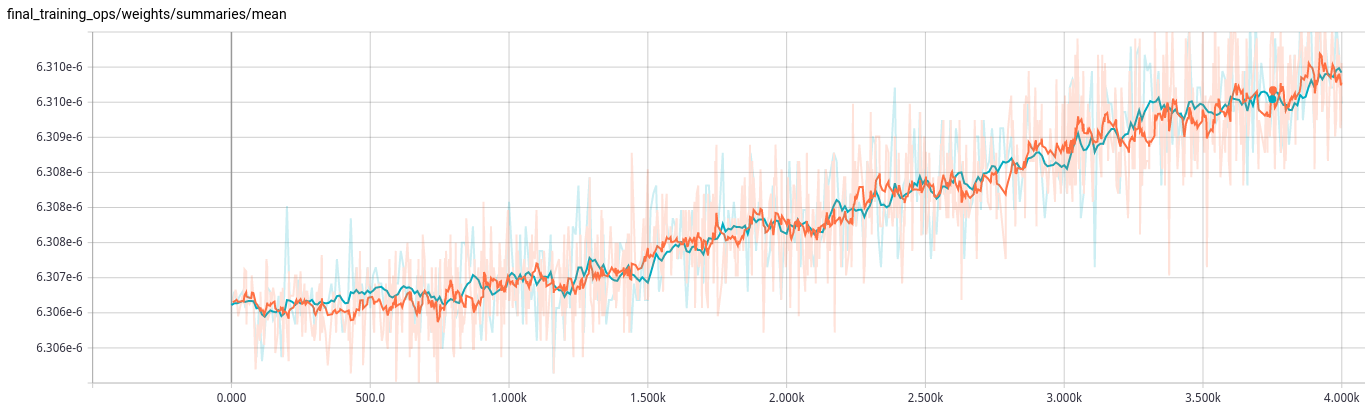}
    \caption[Abstract \acrshort{ai} mean weights graph]{Abstract \acrshort{ai} mean weights graph}
    \label{fig:tensorboard}
\end{figure}

The other way is to use the so called \textit{TensorBoard}, which is a tool contained in the Docker image of TensorFlow. It can produce graphs off the training logs and draws them on a web view so that a researcher can track the progress in real time. Figure \ref{fig:tensorboard} shows a graph visualizing the mean weights of the abstract \acrshort{ai}.

TensorBoard can also create a schematic representation of the \acrlong{nn} (model), which is created during the training. This may not be neccessary in this case, but may become important during the evaluation.

To make the customized Docker image run Tensorboard every time a container is started, the docker-compose.yml has to be modified. The following lines need to be added to the \textit{tensorflow} service in order to do this and make the web view reachable from the host machine:

\begin{minipage}{\linewidth}
\begin{lstlisting}[frame=htrbl, caption={Tensorboard in docker-compose.yml}, label={lst:tensorboard}]
command: tensorboard --logdir /tmp/retrain_logs 
ports:
  - 6006:6006
\end{lstlisting}
\end{minipage}

Because the command defines the process that is run on a image's startup and makes the container exit when it stops, the only possibility to interact with the container after this modification will be \textit{docker exec}. This is not wrong, but rather expected behaviour and should be mentioned at this point.

\subsubsection{Performance}

The aforementioned example to examine an image is a very basic one and does not perform well. By default, it creates a new TensorFlow session for each picture, leading to many unnecessary operations on different parts of the system. To work around this behaviour, the code responsible for the loading of the model and starting the session has to be isolated and put into a function, which is only called once at program startup. Usually, this could be done in the constructor of a class, which represents TensorFlow and would then be passed down in the class it is needed in, but it is not possible to pass arguments to the extended downloader used by the crawler. Thus, the TensorFlow class has to implement a singleton pattern\footnote{\url{http://www.oodesign.com/singleton-pattern.html}}, which means that all instances of this class point to the same object in the \acrshort{ram}. Without the singleton functionality, each downloader spawned by the crawler would work within another TensorFlow session, consuming more resources than actually needed.

\subsubsection{Configuration}

Some variables used by the prototype need to be defined manually and persistently before it is started. In favor of being able to do so, a \acrshort{json} configuration file (listing \ref{lst:config_json}) was created. It is facilitated by a python class that also inherits the singleton pattern, because the configuration has to be read from at several locations in the code and having a single instance of an object is the most efficient choice of implementations for such cases. The class can be used as shown below in listing \ref{lst:config_cls}.

\begin{minipage}{\linewidth}
\begin{lstlisting}[frame=htrbl, caption={Example configuration file}, label={lst:config_json}]
{
    "apiBase": "http://api.gbif.org/v1/species/"
}
\end{lstlisting}
\end{minipage}

\begin{minipage}{\linewidth}
\lstset{language=python}
\begin{lstlisting}[frame=htrbl, caption={Example use of configuration class}, label={lst:config_cls}]
from config import Config

cfg = Config()
print(cfg.get('apiBase'))
#output: 'http://api.gbif.org/v1/species/'
\end{lstlisting}
\end{minipage}

%% file: kapitel/7_evaluation.tex
\section{Evaluation and results} \label{evaluation}

In this chapter, the evaluation process will be explained in-depth by describing a methodical concept for the testing of a dataset. Afterwards, several datasets (including the automatically created one) will be presented and described, so that they can be equitably compared. Finally, the resulting accuracy rate(s) for each dataset will be depcited.

\subsection{Testing concept}

In order to produce comparable results, all datasets must be used to train the \acrshort{ai} using the exact same external preconditions and methodology. This means they have to be brought into the same format and structure, so that all pictures have the same size, e.g. When this was done, the proper testing can be initiated.

There are two procedures that will be used for the evaluation, but both make use of the same principle: Before the training process is started, a subset of the dataset will be isolated and used to determine the error rate of the \acrshort{ai} later on. This needs to be done to avoid overfitting, which is explained in the description of the aforementioned example from TensorFlow\footnote{\url{https://www.tensorflow.org/tutorials/image_retraining\#training_validation_and_testing_sets}}:

\begin{quotation}
\textit{A big potential problem when we're doing machine learning is that our model may just be memorizing irrelevant details of the training images to come up with the right answers. For example, you could imagine a network remembering a pattern in the background of each photo it was shown, and using that to match labels with objects. It could produce good results on all the images it's seen before during training, but then fail on new images because it's not learned general characteristics of the objects, just memorized unimportant details of the training images.}

\textit{This problem is known as overfitting, and to avoid it we keep some of our data out of the training process, so that the model can't memorize them.}
\end{quotation}

In the example's code, this functionality is already built-in. A percentaged accuracy is printed out after every successful training and can be used to estimate the performance of the newly created \acrshort{ai}. This value will be used as a first, approximate result for each dataset. Unfortunately, the example's binary offers no option to keep the isolated images in order to re-use them, but rather puts them back in place without notifying the developer which pictures were isolated.

The problem with this behaviour is that if the datasets have a different amounts of categories, the one with the fewest will probably have the highest accuracy. To resolve this, another test needs to be done. The categories of each dataset have to be sorted out before the training process is started, so that only those existing in either all or at least two datasets are used. For example: Dataset A contains 10 categories, dataset B five and dataset C only three. One run would cover those five categories present in both dataset A and B. The next run then only includes the categories from A and C, and so on. This  makes the evaluation circumstances more even and leads to more comparable results.

\begin{figure}[htbp]
    \centering
    \includegraphics[width=0.55\textwidth]{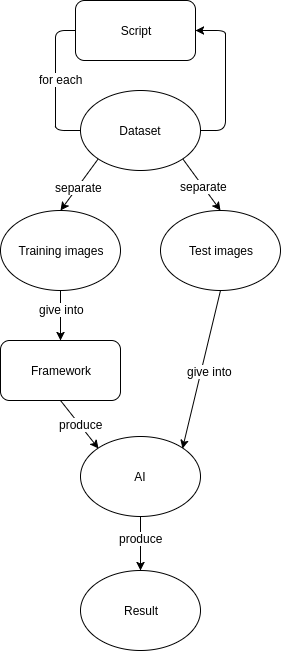}
    \caption[Evaluation process]{Evaluation process}
    \label{fig:evaluation_script}
\end{figure}

The subset for testing will consist of five random images per category picked from the datasets by a rather trivial script. The script is also able to put the images back into the right dataset, so that the test can be repeated with different images and a mean value of the resulting error rates can be found. The evaluation process is visualized in figure \ref{fig:evaluation_script}.

\subsection{Formal dataset comparison}

In 2010, researchers at the California Institute of Technology assembled a dataset containing 200 species of birds. In the following year, they released a revised version called \textit{Caltech-UCSD Birds-200-2011} \cite{cub2011}, which contains more images per category. "\textit{The images were downloaded from the website Flickr and filtered by workers on Amazon Mechanical Turk}" \cite{cub}, as already referenced in \ref{ssec:publications}. This means the dataset was built by laypersons without supervision of biologists or other experts and may contain wrongly assigned entries. Table \ref{table:ds_cub} shows the specifications of this dataset.

\begin{table}[htbp]
\begin{center}
\begin{tabular}{ |l|l| }
    \hline
    \rowcolor[gray]{.9} Attribute & Value \\ \hline
    Size & 1.15 GB \\ \hline
    Pictures & 11788 \\ \hline
    Categories & 14 \\ \hline
    $\varnothing$ Pictures & 842.0 \\ \hline
\end{tabular}
\caption{Caltech-UCSD Birds-200-2011 dataset specifications}
\label{table:ds_cub}
\end{center}
\end{table}

These problems were investigated further in 2015, when a new dataset named \textit{NABirds} was created. In \cite{nabirds}, Caltech-UCSD Birds-200-2011 is picked up and analyzed: "\textit{[...], we found that citizen scientists provide significantly higher quality labels than Mechanical Turk workers, and found that Turkers have alarmingly poor performance annotating fine-grained classes. This has resulted in error rates of over 4\% in fine-grained categories in popular datasets like CUB-200-2011 and ImageNet.}". NABirds is a "\textit{expert-curated dataset}" with an improved performance in comparison to Caltech-UCSD Birds-200-2011, which provides a very high accuracy. Its specifications are listed in table \ref{table:ds_nabirds}.

\begin{table}[htbp]
\begin{center}
\begin{tabular}{ |l|l| }
    \hline
    \rowcolor[gray]{.9} Attribute & Value \\ \hline
    Size & 6.73 GB \\ \hline
    Pictures & 48558 \\ \hline
    Categories & 21 \\ \hline
    $\varnothing$ Pictures & 2312.29 \\ \hline
\end{tabular}
\caption{NABirds dataset specifications}
\label{table:ds_nabirds}
\end{center}
\end{table}

The last dataset to be compared in this evaluation is the one gathered by the prototype. For recapitulation: It was collected (almost) automatically, by letting an abstract \acrshort{ai} do the task of sorting out unwanted pictures, which was done by humans for the two previously presented datasets. The specifications of it can be seen in table \ref{table:ds_auto}.

\begin{table}[htbp]
\begin{center}
\begin{tabular}{ |l|l| }
    \hline
    \rowcolor[gray]{.9} Attribute & Value \\ \hline
    Size & 10.4 GB \\ \hline
    Pictures & 186213 \\ \hline
    Categories & 40 \\ \hline
    $\varnothing$ Pictures & 4655.32 \\ \hline
\end{tabular}
\caption{Automatically created dataset specifications}
\label{table:ds_auto}
\end{center}
\end{table}

If the image numbers of a dataset differs from what is written in the respective paper, this was caused by conversion/resizing errors during the formatting. The number of categories was counted \textit{after} the species were moved to their corresponding order's directory. Also, the given size refers to the reduced size of max. 500 by 500 pixels per image, as described in \ref{sssec:size}.

\subsection{Results}

\begin{figure}[htbp]
    \centering
    \includegraphics[width=1.0\textwidth]{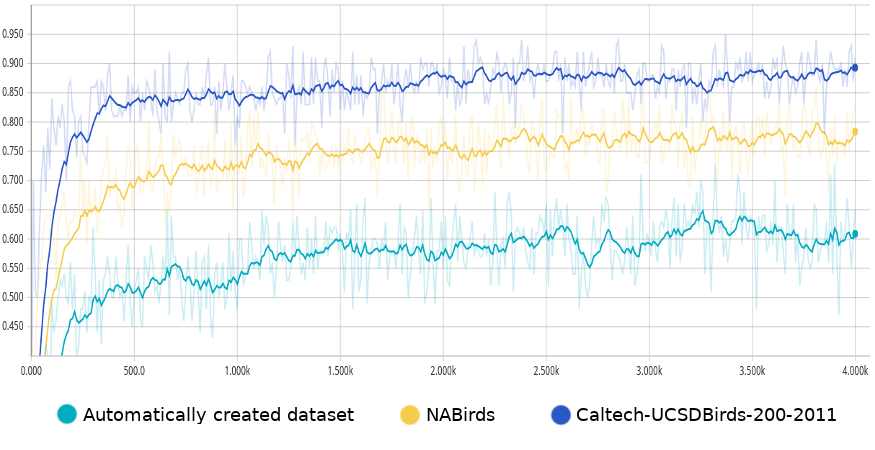}
    \caption[Increase of validation accuracy during training]{Increase of validation accuracy during training}
    \label{fig:tensorflow_results}
\end{figure}

As described before, the first test is built into the TensorFlow retraining example and prints the results after the training. Table \ref{table:results_tensorflow} shows the validation accuracy for each dataset as determined by the example's output. Figure \ref{fig:tensorflow_results} visualizes how the accuracy increased during the training process. The dark blue line represents the Caltech-UCSD Birds-200-2011 dataset, yellow represents NABirds and the turquoise line represents the automatically created dataset.

\begin{table}[htbp]
\begin{center}
\begin{tabular}{ |l|p{3cm}|p{3cm}|p{3cm}| }
    \hline
    \rowcolor[gray]{.9}  & Automated dataset & NABirds & Caltech-UCSD Birds-200-2011 \\ \hline
    Validation accuracy & 63.5\% & 77.8\% & \textbf{85.6}\% \\ \hline
\end{tabular}
\caption{Retraining validation accuracy}
\label{table:results_tensorflow}
\end{center}
\end{table}

These results only represent an estimation for the real-word performance of the \acrshort{ai}, as mentioned in the last subsection. But still, they seem to point out that more categories lead to a worse result in the context of image classification, which illustrates the importance of the following test step.

Each run was done five times in order to get a meaningful average accuracy value. In the result tables, the best outcomes are highlighted using a \textbf{bold} font. If a dataset is marked with an asterisk (*), this means that it misses categories present in the specific run. For example: The result of the first run (see table \ref{table:results_auto}) is pretty obvious, because it was done using all categories from the automated dataset, so that NABirds and Caltech-UCSD Birds-200-2011 miss about half of them and thus can not show a good performance. 

\begin{table}[htbp]
\begin{center}
\begin{tabular}{ |l|p{3cm}|p{3cm}|p{3cm}| }
    \hline
    \rowcolor[gray]{.9} Accuracy & Automated dataset & NABirds* & Caltech-UCSD Birds-200-2011* \\ \hline
    Top-1 & \textbf{60.65}\% & 48.81\% & 39.19\% \\ \hline
    Top-5 & \textbf{82.38}\% & 67.24\% & 52.59\% \\ \hline
\end{tabular}
\caption{Results of run with categories from automated dataset}
\label{table:results_auto}
\end{center}
\end{table}

In the second run (see table \ref{table:results_nabirds}), the automated dataset was reduced to only contain the categories present in NABirds in order to produce a comparable result. The Caltech-UCSD Birds-200-2011 dataset is missing several categories again and was only measured for reference.

\begin{table}[htbp]
\begin{center}
\begin{tabular}{ |l|p{3cm}|p{3cm}|p{3cm}| }
    \hline
    \rowcolor[gray]{.9} Accuracy & Automated dataset & NABirds & Caltech-UCSD Birds-200-2011* \\ \hline
    Top-1 & 68.82\% & \textbf{69.03}\% & 51.85\% \\ \hline
    Top-5 & \textbf{92.44}\% & \textbf{92.44}\% & 70.3\% \\ \hline
\end{tabular}
\caption{Results of run with categories from NABirds}
\label{table:results_nabirds}
\end{center}
\end{table}

The third run was done using only the 14 categories present in Caltech-UCSD Birds-200-2011, meaning the automated dataset had to be reduced once again and NABirds was missing one category. But as the results in table \ref{table:results_cub} show, NABirds performed only slightly worse than the other two datasets.

\begin{table}[htbp]
\begin{center}
\begin{tabular}{ |l|p{3cm}|p{3cm}|p{3cm}| }
    \hline
    \rowcolor[gray]{.9} Accuracy & Automated dataset & NABirds* & Caltech-UCSD Birds-200-2011 \\ \hline
    Top-1 & \textbf{74.93}\% & 71.8\% & 71.9\% \\ \hline
    Top-5 & \textbf{96.2}\% & 92.29\% & 95.41\% \\ \hline
\end{tabular}
\caption{Results of run with categories from Auto}
\label{table:results_cub}
\end{center}
\end{table}

Finally, only the categories present in \textit{all} datasets were considered in the test run, making it the one with the fewest. But surprisingly, the best overall score for both the top-1 and top-5 accuracy was reached in the prior run.

\begin{table}[htbp]
\begin{center}
\begin{tabular}{ |l|p{3cm}|p{3cm}|p{3cm}| }
    \hline
    \rowcolor[gray]{.9} Accuracy & Automated dataset & NABirds & Caltech-UCSD Birds-200-2011 \\ \hline
    Top-1 & 72.4\% & \textbf{73.0}\% & 69.9\% \\ \hline
    Top-5 & \textbf{95.7}\% & 93.5\% & 92.6\% \\ \hline
\end{tabular}
\caption{Results of run with categories present in all datasets}
\label{table:results_all}
\end{center}
\end{table}

\begin{figure}[htbp]
    \centering
    \includegraphics[width=1.0\textwidth]{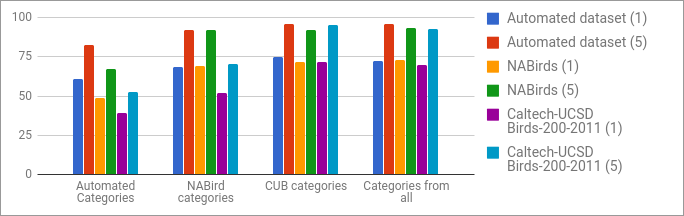}
    \caption[Evaluation results diagram]{Evaluation results diagram}
    \label{fig:eval_results}
\end{figure}

\newpage
All results are visualized again in figure \ref{fig:eval_results} for a better overview. It shows that the automated dataset could reach the best accuracy in most cases.

%% file: kapitel/8_conclusion.tex
\section{Conclusion, reflection and future prospects}\label{conclusion}

At this point, the thesis shall be concluded, the work reflected and the results of the evaluation discussed. Furthermore, ideas of how the prototype could be improved and used efficiently in productive systems are presented.

\subsection{Conclusion}

Within the scope of this thesis, the aim was to find out wether it is practicable to automate the creation of a dataset for image classification or not. To achieve this, the topics of \acrlong{ml} and taxonomy were studied and described in theory. Afterwards, the requirements for a prototype were defined and a concept for it was presented. As a practical part of the work, a proof-of-concept system was prototypically implemented and the outcoming, automatically created dataset was evaluated.

During the implementation, several unforeseen problems surfaced, so that the proposed system had to be modified at various points in order to tackle them. Also, the scope of classification had to be reduced from species to orders, because detecting the former was too ambitious. Nonetheless, the evaluation showed a very good result.

One thing that became visible after the evaluation is that "\textit{Quantity can be more important than quality}", which is also described in \cite{nabirds}. The NABirds dataset may contain images of better quality with fewer false labels than the automatically created one, but in the context of \acrlong{ml}, more data often leads to better results (as stated already in chapter \ref{introduction}) and computers that collect and sort (digital) information can do this much faster than their human counterparts.

Of course, there also is a drawback in the approach to automate the dataset creation proposed in this work; it can only be applied to abstractable categories, as mentioned in \ref{sssec:abstract_ai}. While it would work perfectly with different types of balls (tennis balls, footballs, etc.), it would make no sense when trying to gather a dataset for categories that do not share (abstract) attributes, like the ones included in the ImageNet.

In conclusion, it can be stated that the key questions of this thesis can be answered explicitly positively and the prototypical implementation can be used as a solid base for the integration of an \acrlong{ai} which is able to detect species into one of the systems inside the biodiversity warehouse.

\subsection{Future prospects}

The implementation works well for a prototype, but may be optimized in means of performance when being integrated into a productive system. As written in \ref{sssec:hardware}, one of the parts limiting the speed during the training of an \acrlong{ai} is the \acrshort{ssd}/\acrshort{hdd}. To bypass the "slow" reading and writing operations of such a device, one could use a so-called \textit{ramdisk}. Ramdisks are virtual hard drives that save their data inside the system's \acrshort{ram}. They outperform all currently available \acrshortpl{ssd}\footnote{\url{https://www.raymond.cc/blog/12-ram-disk-software-benchmarked-for-fastest-read-and-write-speed/}} and can be created using freely available software. The only thing to consider when thinking about using ramdisks in this case is the overall size of the memory, because a ramdisk will decrease it by the factor of its own space and there must be left enough to hold the redis \acrlong{db}.

If the approach of this thesis is adapted and employed inside any kind of application, there could be a way to achieve the detection of species after all. Because taxonomy follows a strict hierarchy, it might be possible to concatenate different layers containing several \acrlongpl{ai}, so that each of them would represent one taxnomic layer. If an image is given into the \acrshort{ai}-tree, it could be analyzed and redirected to the resulting child-\acrshort{ai}. 

For example, if the picture shows a duck, it is classified as a bird (class) by the first \acrshort{ai}, then as a member of Anseriformes (order) by the second. Afterwards, it is detected as belonging to Anatidae (family) by the \acrlong{ai} responsible for determining the families under the order Anseriformes. It will then be given into the \acrlong{ai} for Anatidae and be defined as being associated with Anas (genus). Finally, the \acrlong{ai} for classifying member of the genus Anas will determine its species (Anas platyrhynchos).

The forestanding example is visualized in figure \ref{fig:ai_tree} (see \textit{Attachments}). Making use of this idea would also solve the problem of having to train an \acrlong{ai} with too many categories for lower taxa, hence increasing the detection rate.

%% file: extras/anhang.tex
\addcontentsline{toc}{section}{Attachments}
\fancyhead[L]{Attachments}
\section*{Attachments}\label{anhang}

\begin{figure}[htbp]
    \centering
    \includegraphics[width=0.925\textwidth]{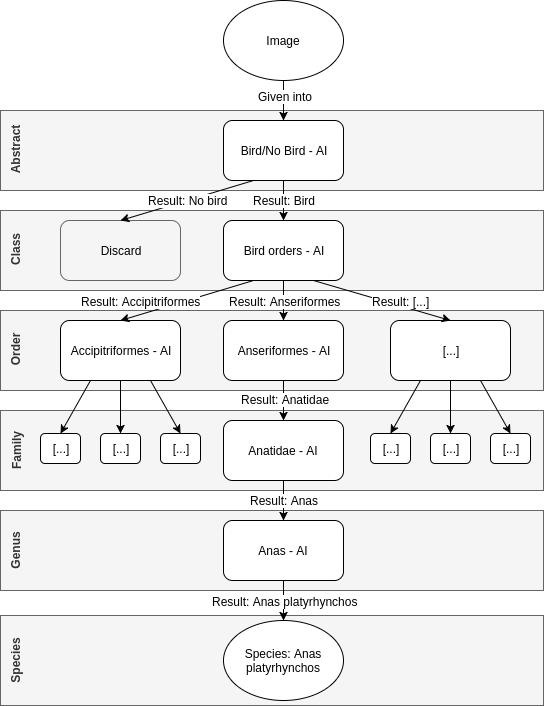}
    \caption[Visualization of an \acrshort{ai}-tree]{Visualization of an \acrshort{ai}-tree}
    \label{fig:ai_tree}
\end{figure}

%% file: output.bbl
\newcommand{\etalchar}[1]{$^{#1}$}
\begin{thebibliography}{WBW{\etalchar{+}}11}

\bibitem[AAB{\etalchar{+}}15]{tensorflow}
Mart\'{\i}n Abadi, Ashish Agarwal, Paul Barham, Eugene Brevdo, Zhifeng Chen,
  Craig Citro, Greg~S. Corrado, Andy Davis, Jeffrey Dean, Matthieu Devin,
  Sanjay Ghemawat, Ian Goodfellow, Andrew Harp, Geoffrey Irving, Michael Isard,
  Yangqing Jia, Rafal Jozefowicz, Lukasz Kaiser, Manjunath Kudlur, Josh
  Levenberg, Dan Man\'{e}, Rajat Monga, Sherry Moore, Derek Murray, Chris Olah,
  Mike Schuster, Jonathon Shlens, Benoit Steiner, Ilya Sutskever, Kunal Talwar,
  Paul Tucker, Vincent Vanhoucke, Vijay Vasudevan, Fernanda Vi\'{e}gas, Oriol
  Vinyals, Pete Warden, Martin Wattenberg, Martin Wicke, Yuan Yu, and Xiaoqiang
  Zheng.
\newblock {TensorFlow}: Large-scale machine learning on heterogeneous systems,
  2015.
\newblock Software available from tensorflow.org.

\bibitem[AONA10]{fishrec}
Mutasem~Khalil Alsmadi, Khairuddin~Bin Omar, Shahrul~Azman Noah, and Ibrahim
  Almarashdeh.
\newblock Fish recognition based on robust features extraction from size and
  shape measurements using neural network.
\newblock {\em Journal of Computer Science}, 6, 2010.

\bibitem[BL89]{html}
Tim Berners-Lee.
\newblock Information management: A proposal.
\newblock Technical report, CERN, 1989.

\bibitem[BRSS15]{compstudy}
Soheil Bahrampour, Naveen Ramakrishnan, Lukas Schott, and Mohak Shah.
\newblock Comparative study of caffe, neon, theano, and torch for deep
  learning.
\newblock {\em CoRR}, abs/1511.06435, 2015.

\bibitem[Cai17]{tax1}
A.J. Cain.
\newblock Taxonomy.
\newblock {\em Encyclopaedia Britannica}, 2017.
\newblock \url{https://www.britannica.com/science/taxonomy}.

\bibitem[Cas15]{amzndata}
Alex Casalboni.
\newblock Amazon mechanical turk: help for building your machine learning
  datasets.
\newblock Blogpost, 2015.
\newblock
  \url{https://cloudacademy.com/blog/machine-learning-datasets-mechanical-turk/}.

\bibitem[CKF11]{torch7}
Ronan Collobert, Koray Kavukcuoglu, and Cl{\'{e}}ment Farabet.
\newblock Torch7: A matlab-like environment for machine learning.
\newblock In {\em BigLearn, NIPS Workshop}, 2011.

\bibitem[CS98]{tax3}
Thomas Cavalier-Smith.
\newblock A revised six-kingdom system of life.
\newblock {\em Biological Reviews}, 73(3):203--266, 1998.

\bibitem[DRLF17]{watermarks}
Tali Dekel, Michael Rubinstein, Ce~Liu, and William~T. Freeman.
\newblock On the effectiveness of visible watermarks.
\newblock Technical report, Google Research, 2017.

\bibitem[FFLR{\etalchar{+}}16]{imagenet}
Li~Fei-Fei, Kai Li, Olga Russakovsky, Jia~Deng Jonathan~Krause, and Alex Berg.
\newblock Imagenet.
\newblock Website, 2016.
\newblock \url{http://www.image-net.org/}.

\bibitem[Fri17]{fritsche}
Jannis Fritsche.
\newblock {Gamification von Biodiverstäts Atlanten am Beispiel vom
  Säugetieratlas Bremen.} ({German}) [{Gamification of biodiversity atlases
  using the example of the Mammals Atlas Bremen}].
\newblock Bachelor's thesis, Hochschule Bremen - City University of Applied
  Sciences, 2017.

\bibitem[GPC{\etalchar{+}}16]{health2}
Varun Gulshan, Lily Peng, Marc Coram, Martin~C. Stumpe, Derek Wu, Arunachalam
  Narayanaswamy, Subhashini Venugopalan, Kasumi Widner, Tom Madams, Jorge
  Cuadros, Ramasamy Kim, Rajiv Raman, Philip~C. Nelson, Jessica~L. Mega, and
  Dale~R. Webster.
\newblock Development and validation of a deep learning algorithm for detection
  of diabetic retinopathy in retinal fundus photographs.
\newblock {\em JAMA}, 316(22):2402--2410, 2016.

\bibitem[Gro16]{grossman}
D.~Grossman.
\newblock 850k images in 24 hours: Automating deep learning dataset creation.
\newblock Blogpost, 2016.
\newblock
  \url{https://gab41.lab41.org/850k-images-in-24-hours-automating-deep-learning-dataset-creation-60bdced04275}.

\bibitem[HBF{\etalchar{+}}15]{nabirds}
G.~Van Horn, S.~Branson, R.~Farrell, S.~Haber, J.~Barry, P.~Ipeirotis,
  P.~Perona, and S.~Belongie.
\newblock Building a bird recognition app and large scale dataset with citizen
  scientists: The fine print in fine-grained dataset collection.
\newblock In {\em 2015 IEEE Conference on Computer Vision and Pattern
  Recognition (CVPR)}, pages 595--604, June 2015.

\bibitem[Hei17]{heil}
Raphaela~Marion Heil.
\newblock {Automatic Validation of Biodiversity Data}.
\newblock Master's thesis, Hochschule Bremen - City University of Applied
  Sciences, 2017.

\bibitem[Hel11]{mil1}
Arnie Heller.
\newblock From video to knowledge.
\newblock {\em Science \& Technology Review}, April/May 2011.
\newblock \url{https://str.llnl.gov/AprMay11/vaidya.html}.

\bibitem[Hen66]{tax2}
Willi Hennig.
\newblock {\em Phylogenetic systematics}.
\newblock Staatliches Museum für Naturkunde in Stuttgart, Germany, 1966.

\bibitem[JBG{\etalchar{+}}16]{plantnet}
Alexis Joly, Pierre Bonnet, Herv{\'e} Go{\"e}au, Julien Barbe, Souheil Selmi,
  Julien Champ, Samuel Dufour-Kowalski, Antoine Affouard, Jennifer Carr{\'e},
  Jean-Fran{\c{c}}ois Molino, Nozha Boujemaa, and Daniel Barth{\'e}l{\'e}my.
\newblock A look inside the pl@ntnet experience.
\newblock {\em Multimedia Systems}, 22(6):751--766, Nov 2016.

\bibitem[Jeg17]{jeg}
Fabian Jeglinski.
\newblock {Realisierung eines digitalen Werkzeugs zur Erstellung und Verwaltung
  von Descriptive Data im Kontext der taxonomischen Bestimmung von Lebewesen}
  ({German}) [{Realisation of a digital tool for the creation and management of
  descriptive data in the context of taxonomic creature identification}].
\newblock Master's thesis, Hochschule Bremen - City University of Applied
  Sciences, 2017.

\bibitem[JSD{\etalchar{+}}14]{caffe}
Yangqing Jia, Evan Shelhamer, Jeff Donahue, Sergey Karayev, Jonathan Long, Ross
  Girshick, Sergio Guadarrama, and Trevor Darrell.
\newblock Caffe: Convolutional architecture for fast feature embedding.
\newblock In {\em Proceedings of the 22Nd ACM International Conference on
  Multimedia}, MM '14, pages 675--678, New York, NY, USA, 2014. ACM.

\bibitem[KKK16]{speedacc}
Vassili Kovalev, Alexander Kalinovsky, and Sergey Kovalev.
\newblock Deep learning with theano, torch, caffe, tensorflow, and
  deeplearning4j: Which one is the best in speed and accuracy?
\newblock Technical report, United Institute of Informatics Problems, Belarus
  National Academy of Sciences, 2016.

\bibitem[Kra11]{avghash}
Neal Krawetz.
\newblock Looks like it.
\newblock Blogpost, 2011.
\newblock
  \url{http://www.hackerfactor.com/blog/index.php?/archives/432-Looks-Like-It.html}.

\bibitem[KS14]{vgg}
Andrew~Zisserman Karen~Simonyan.
\newblock Very deep convolutional networks for large-scale image recognition.
\newblock Technical report, Visual Geometry Group, Department of Engineering
  Science, University of Oxford, 2014.

\bibitem[KSH01]{som}
T.~Kohonen, M.~R. Schroeder, and T.~S. Huang, editors.
\newblock {\em Self-Organizing Maps}.
\newblock Springer-Verlag New York, Inc., Secaucus, NJ, USA, 3rd edition, 2001.

\bibitem[KSH12a]{deep1}
Alex Krizhevsky, Ilya Sutskever, and Geoffrey~E. Hinton.
\newblock Imagenet classification with deep convolutional neural networks.
\newblock Technical report, University of Toronto, 2012.

\bibitem[KSH12b]{alexnet}
Alex Krizhevsky, Ilya Sutskever, and Geoffrey~E Hinton.
\newblock Imagenet classification with deep convolutional neural networks.
\newblock In F.~Pereira, C.~J.~C. Burges, L.~Bottou, and K.~Q. Weinberger,
  editors, {\em Advances in Neural Information Processing Systems 25}, pages
  1097--1105. Curran Associates, Inc., 2012.

\bibitem[KWKT15]{dcign}
Tejas~D. Kulkarni, Will Whitney, Pushmeet Kohli, and Joshua~B. Tenenbaum.
\newblock Deep convolutional inverse graphics network.
\newblock Technical report, Computer Science and Artificial Intelligence
  Laboratory (MIT), Brain and Cognitive Sciences (MIT) and Microsoft Research
  Cambridge, 2015.

\bibitem[LGN{\etalchar{+}}17]{health1}
Yun Liu, Krishna Gadepalli, Mohammad Norouzi, George~E. Dahl, Timo Kohlberger,
  Aleksey Boyko, Subhashini Venugopalan, Aleksei Timofeev, Philip~Q. Nelson,
  Greg~S. Corrado, Jason~D. Hipp, Lily Peng, and Martin~C. Stumpe.
\newblock Detecting cancer metastases on gigapixel pathology images.
\newblock Technical report, Google Brain, Google Inc, Verily Life Sciences,
  2017.

\bibitem[LZ17]{mlnn}
Quoc Le and Barret Zoph.
\newblock Using machine learning to explore neural network architecture.
\newblock Blogpost, 2017.
\newblock
  \url{https://research.googleblog.com/2017/05/using-machine-learning-to-explore.html}.

\bibitem[MBC10]{nature}
Norman MacLeod, Mark Benfield, and Phil Culverhouse.
\newblock Time to automate identification.
\newblock {\em nature}, 467:154--155, 2010.

\bibitem[NLM{\etalchar{+}}12]{deep2}
Andrew~Y. Ng, Quoc~V. Le, Marc’Aurelio, Rajat Monga, Matthieu Devin, Kai
  Chen, Greg~S. Corrado, and Jeff Dean.
\newblock Building high-level features using large scale unsupervised learning.
\newblock In {\em Proceedings of the 29 th International Conference on Machine
  Learning}, 2012.

\bibitem[OMS10]{mil3}
Omar Oreifej, Ramin Mehran, and Mubarak Shah.
\newblock Human identity recognition in aerial images.
\newblock Technical report, Computer Vision Lab, University of Central Florida,
  2010.

\bibitem[O'N10]{daisy}
Mark~A. O'Neill.
\newblock Daisy: A practical tool for automated species identification.
\newblock Technical report, Tumbling Dice Ltd., University of Newcastle upon
  Tyne, 2010.

\bibitem[PKS07]{neuroscience}
Steven~M. Platek, Julian~Paul Keenan, and Todd~K. Shackelford, editors.
\newblock {\em Evolutionary Cognitive Neuroscience}.
\newblock Massachusetts Institute of Technology, 2007.

\bibitem[PZM12]{mil2}
Jan Prokaj, Xuemei Zhao, and Gerard Medioni.
\newblock Tracking many vehicles in wide area aerial surveillance.
\newblock Technical report, University of Southern California, 2012.

\bibitem[RDB16]{art}
Manuel Ruder, Alexey Dosovitskiy, and Thomas Brox.
\newblock Artistic style transfer for videos.
\newblock Technical report, Department of Computer Science, University of
  Freiburg, 2016.

\bibitem[Sch17]{schrader}
Jan~Christoph Schrader.
\newblock {Konzeption, prototypische Entwicklung und Evaluation eines Systems
  zur Ermittlung von Eigenschaften von Spezies mittels Natural Language
  Processing} ({German}) [{Conception, prototypical development and evaluation
  of a system to detect attributes of species with natural language
  processing}].
\newblock Master's thesis, Hochschule Bremen - City University of Applied
  Sciences, 2017.

\bibitem[SDAM07]{tax4}
Edna Suárez-Díaz and Victor~H. Anaya-Muñoz.
\newblock History, objectivity, and the construction of molecular phylogenies.
\newblock {\em Studies in History and Philosophy of Science Part C: Studies in
  History and Philosophy of Biological and Biomedical Sciences}, 2007.

\bibitem[SEZ{\etalchar{+}}13]{overfeat}
Pierre Sermanet, David Eigen, Xiang Zhang, Michael Mathieu, Rob Fergus, and
  Yann LeCun.
\newblock Overfeat: Integrated recognition, localization and detection using
  convolutional networks.
\newblock Technical report, Courant Institute of Mathematical Sciences, New
  York University, 2013.

\bibitem[SLJ{\etalchar{+}}15]{googlenet}
C.~Szegedy, Wei Liu, Yangqing Jia, P.~Sermanet, S.~Reed, D.~Anguelov, D.~Erhan,
  V.~Vanhoucke, and A.~Rabinovich.
\newblock Going deeper with convolutions.
\newblock In {\em 2015 IEEE Conference on Computer Vision and Pattern
  Recognition (CVPR)}, pages 1--9, June 2015.

\bibitem[Sok17]{sec}
Daniel~AJ Sokolov.
\newblock {Autonome Systeme und Künstliche Intelligenz: Maschinen und Hacker
  kooperieren bei Hacking-Wettbewerben} ({German}) [{Autonomous systems and
  artificial intelligence: Machines and hackers cooperate at hacking
  contests}].
\newblock News article, February 2017.
\newblock
  \url{https://www.heise.de/security/meldung/Autonome-Systeme-und-Kuenstliche-Intelligenz-Maschinen-und-Hacker-kooperieren-bei-Hacking-3632244.html}.

\bibitem[Ste10]{fin2}
Christopher Steiner.
\newblock Wall street's speed war.
\newblock News article, September 2010.
\newblock
  \url{https://www.forbes.com/forbes/2010/0927/outfront-netscape-jim-barksdale-daniel-spivey-wall-street-speed-war.html}.

\bibitem[SVI{\etalchar{+}}15]{inceptionv3}
Christian Szegedy, Vincent Vanhoucke, Sergey Ioffe, Jonathon Shlens, and
  Zbigniew Wojna.
\newblock Rethinking the inception architecture for computer vision.
\newblock {\em CoRR}, abs/1512.00567, 2015.

\bibitem[{The}16]{theano}
{Theano Development Team}.
\newblock {Theano: A {Python} framework for fast computation of mathematical
  expressions}.
\newblock {\em arXiv e-prints}, abs/1605.02688, May 2016.

\bibitem[Tra15]{trask}
Andrew Trask.
\newblock A neural network in 11 lines of python.
\newblock Blogpost, 2015.
\newblock \url{http://iamtrask.github.io/2015/07/12/basic-python-network/}.

\bibitem[TW09]{fin}
C.-F. Tsai and S.-P. Wang.
\newblock Stock price forecasting by hybrid machine learning techniques.
\newblock {\em Proceedings of the International MultiConference of Engineers
  and Computer Scientists}, 1, 2009.

\bibitem[vV16]{veen}
Fjodor van Veen.
\newblock The neural network zoo.
\newblock Blogpost, 2016.
\newblock \url{http://www.asimovinstitute.org/neural-network-zoo/}.

\bibitem[WBM{\etalchar{+}}10]{cub}
P.~Welinder, S.~Branson, T.~Mita, C.~Wah, F.~Schroff, S.~Belongie, and
  P.~Perona.
\newblock {Caltech-UCSD Birds 200}.
\newblock Technical Report CNS-TR-2010-001, California Institute of Technology,
  2010.

\bibitem[WBW{\etalchar{+}}11]{cub2011}
C.~Wah, S.~Branson, P.~Welinder, P.~Perona, and S.~Belongie.
\newblock {The Caltech-UCSD Birds-200-2011 Dataset}.
\newblock Technical Report CNS-TR-2011-001, California Institute of Technology,
  2011.

\bibitem[ZH17]{health3}
Ping Zhang and Jianying Hu.
\newblock Featured patent: Machine learning models for drug discovery.
\newblock Blogpost, 2017.
\newblock
  \url{https://www.ibm.com/blogs/research/2017/04/machine-learning-models-drug-discovery/}.

\bibitem[ZZI{\etalchar{+}}17]{art2}
Richard Zhang, Jun-Yan Zhu, Phillip Isola, Xinyang Geng, Angela~S. Lin, Tianhe
  Yu, and Alexei~A. Efros.
\newblock Real-time user-guided image colorization with learned deep priors.
\newblock Technical report, University of California, Berkeley, 2017.

\end{thebibliography}
